\documentclass[runningheads]{llncs}

\usepackage{eccv}

\usepackage{eccvabbrv}

\usepackage{graphicx}
\usepackage{booktabs}

\usepackage[accsupp]{axessibility}  

\usepackage{hyperref}

\usepackage{orcidlink}
\usepackage{amsfonts}
\usepackage{array}
\usepackage{bbm}
\usepackage{booktabs}
\usepackage{enumitem}
\usepackage{float}
\usepackage[symbol]{footmisc}
\usepackage{mathtools}
\usepackage{mathrsfs}
\usepackage{multirow}
\usepackage{setspace}
\usepackage{subcaption}
\usepackage{xfrac}
\usepackage{pifont}
\usepackage{amsmath}
\usepackage[normalem]{ulem}
\usepackage{xcolor}
\usepackage{color, colortbl}

\DeclareUnicodeCharacter{3000}{  }
\DeclareUnicodeCharacter{2212}{-}
\newcommand{\cmark}{\textcolor{green!80!black}{\ding{51}}}
\newcommand{\xmark}{\textcolor{red}{\ding{55}}}

\newcolumntype{L}[1]{>{\raggedright\let\newline\\\arraybackslash\hspace{0pt}}m{#1}}
\newcolumntype{C}[1]{>{\centering\let\newline\\\arraybackslash\hspace{0pt}}m{#1}}
\newcolumntype{R}[1]{>{\raggedleft\let\newline\\\arraybackslash\hspace{0pt}}m{#1}}

\setlist[itemize]{noitemsep, topsep=0pt}
\setlist[enumerate]{noitemsep, topsep=0pt}

\newcommand{\parens}[1]{\left(#1\right)}
\newcommand{\braces}[1]{\left\{#1\right\}}
\newcommand{\bracks}[1]{\left[#1\right]}
\newcommand{\modulus}[1]{\left\vert#1\right\vert}
\newcommand{\norm}[1]{\left\Vert#1\right\Vert}

\DeclareMathOperator*{\argmin}{arg\,min}

\newcommand{\spsparens}[1]{^{\parens{#1}}}

\definecolor{Gray}{gray}{0.9}

\newcommand{\method}{ReMoS}
\newcommand{\dataset}{ReMoCap}
\newcommand{\bodyatt}{CoST-XA}
\newcommand{\handatt}{H-XA}
\newcommand{\gpu}{an NVIDIA RTX A4000 GPU}

\newcommand{\textitblue}{\textcolor{blue}{\textit{blue}}}
\newcommand{\textitred}{\textcolor{red}{\textit{red}}}

\definecolor{RDcolor}{rgb}{0.0, 0.8, 0.75}

\newcommand{\bulletitem}{\item[$\bullet$]}

\definecolor{AGcolor}{rgb}{0.0, 0.1, 1.0}

\definecolor{darkorange}{rgb}{1.0, 0.55, 0.0}

\definecolor{modification_color}{rgb}{1.0, 0.1, 0.4}

\begin{document}

\title{\textsc{\method}: 3D Motion-Conditioned Reaction Synthesis for Two-Person Interactions}
\titlerunning{ReMoS}

\author{Anindita Ghosh\inst{1,2}\orcidlink{0000-0001-5361-8806} \and
Rishabh Dabral\inst{2}\orcidlink{0009-0004-1245-4146} 
\and
Vladislav Golyanik\inst{2}\orcidlink{0000-0003-1630-2006} 
\and
Christian Theobalt\inst{2}\orcidlink{0000-0001-6104-6625} 
\and
Philipp Slusallek\inst{1,2}\orcidlink{0000-0002-2189-2429}}

\authorrunning{A. Ghosh et al.}


\institute{German Research Center for Artificial Intelligence (DFKI)
\and
Max-Planck Institute for Informatics (MPII)
 \and
Saarland Informatics Campus
}

\maketitle

\begin{abstract}
    Current approaches for 3D human motion synthesis generate high-quality animations of digital humans performing a wide variety of actions and gestures.
    However, a notable technological gap exists in addressing the complex dynamics of multi-human interactions within this paradigm.
    In this work, we present \method, a denoising diffusion-based model that synthesizes full-body \textit{reactive motion} of a person in a two-person interaction scenario. 
    Given the motion of one person, 
    we employ a combined spatio-temporal cross-attention mechanism
    to synthesize the reactive body and hand motion of the second person, thereby completing the interactions between the two.
    We demonstrate \method~across challenging two-person scenarios such as pair-dancing, Ninjutsu, kickboxing, and acrobatics, where one person's movements have complex and diverse influences on the other.
    We also contribute the \dataset~dataset for two-person interactions containing full-body and finger motions.
    We evaluate \method~through multiple quantitative metrics, qualitative visualizations, and a user study,
    and also indicate usability in interactive motion editing applications.
    More details are available on the project page: \href{https://vcai.mpi-inf.mpg.de/projects/remos/}{https://vcai.mpi-inf.mpg.de/projects/remos}.
    
\end{abstract}

\section{Introduction}
\label{sec:intro}

Digital 3D character motion synthesis 
has emerged as the next frontier for animation pipelines~\cite{PoYifanGolyanik2023DiffSTAR},
particularly through denoising diffusion probabilistic models (DDPMs)~\cite{SohlDickstein2015}.
While methods for generating character motion for various tasks
such as text or music conditioned motion synthesis~\cite{ghosh2021text, petrovich2022temos, athanasiou2022teach, aristidou2022rhythm, huang2022genre, guo2022tm2t, guo2022generating, zhang2023generating}, face and gesture synthesis~\cite{ng2022learning, xing2023codetalker, bhattacharya2021text2gestures, bhattacharya2021speech2affectivegestures, habibie2022motion, Mughal_2024_CVPR}, human-scene interaction~\cite{ghosh2022imos, zhang2022couch, Zhang2023roam, wang2024move} exist, 
synthesizing interactions \textit{between} humans is relatively under-explored.
Modeling such human-human interactions is essential for designing generative 3D human motion synthesis frameworks supporting the complex physical and social interplay of two interacting persons~\cite{tanke2023social}. It offers new capabilities for character animation tools and software,
 with applications in commercial and entertainment media~\cite{hanser2009scenemaker}, interactive mixed and augmented reality~\cite{egges2007presence}, and social robotics~\cite{yoon2019robots}.
\begin{figure}[t]
    \centering
    \includegraphics[width=1.0\linewidth]{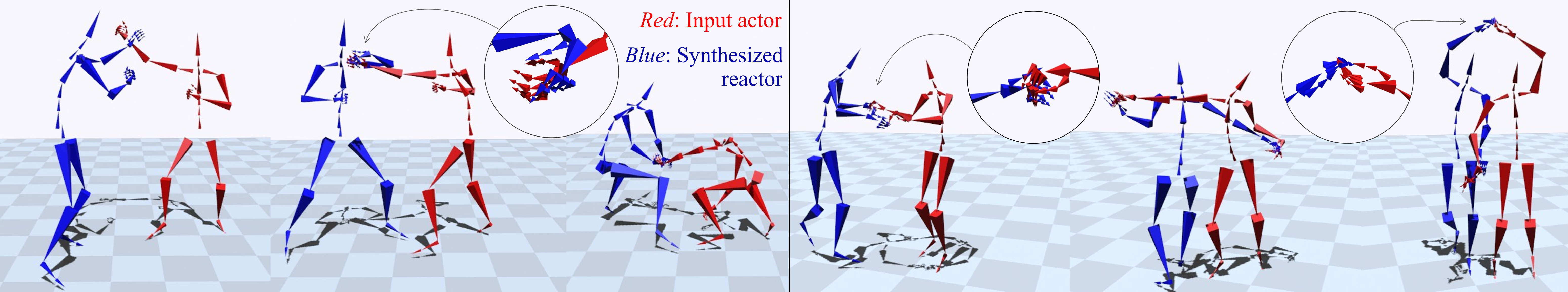}
    \caption{\textbf{Visualizations of reactive 3D motion sequences synthesized with the proposed \method~approach.} We synthesize the 3D full-body motion of the reactor (\textitblue) conditioned only on the 3D motion of the actor (\textitred), thereby completing the interactions between the two (Ninjutsu practice on the \textit{left} and Lindy Hop dancing on the \textit{right}). 
    The synthesized hand interactions are enlarged and highlighted with \textit{circles}. 
    }
    \label{fig:teaser} 
\end{figure} 
\par
Toward this goal, we focus on the task of modeling reactive motions.
We aim to automatically generate realistic, temporally-aligned reactive motions of a responding person given the continuous motion of a guiding person 
(see Fig.~\ref{fig:teaser}). 
This alleviates a common overhead for animators, enabling them to design an \textit{acting} character and automatically obtain meaningful motions for a \textit{reacting} character.
However, automating such a process poses two key challenges. 
First, two-person interaction synthesis increases the dimensionality of the already challenging problem of single-person motion synthesis. 
In our case, the generated reactive motions must align with the conditional signals provided in the form of 3D motion sequences from the actor.
Second, doing so in a generative setting without any cues from text prompts or action labels (to obviate additional supervision and data needs) requires a careful trade-off between generating diverse motions and adhering to the narrow manifold of plausible reactions.
While existing methods for synthesizing interactions~\cite{goel2022interaction, chopin2023bipartite} and two-person motions~\cite{tanaka2023role, liang2023intergen, shafir2023human} are good at generating plausible motions, they rely on additional annotations, such as action labels or text prompts, to specify the motion.
Action labels typically depict actions at a high level, such as ``high-five'' or ``salsa'', but do not capture the fine-grained synchronization with the actor's motions.
Textual descriptions offer nuance for the motions, but collecting accurate textual annotations is considerably cumbersome, particularly for contact-heavy or fast-paced two-person motions, such as dancing and exercising.
Further, performing generative tasks through textual prompting remains challenging for non-experts~\cite{why_johnny_cant_prompt}.
Apart from the data, contact-based motion generation also needs to consider hand-based interactions to improve realism.
This entails the need for frameworks capable of synthesizing hand movements with higher degrees of freedom, which are challenging to model but crucial for improving motion realism.
Balancing focus between the broader \textit{full-body reactive movements} and the finer \textit{hand motions} involves addressing motion at two significantly different scales --- a challenge exacerbated by the 
high degrees of freedom of the skeletons.
\par
\begin{figure}[t]
    \begin{minipage}[c]{0.5\linewidth}
    \centering
        \includegraphics[width=1.0\linewidth]{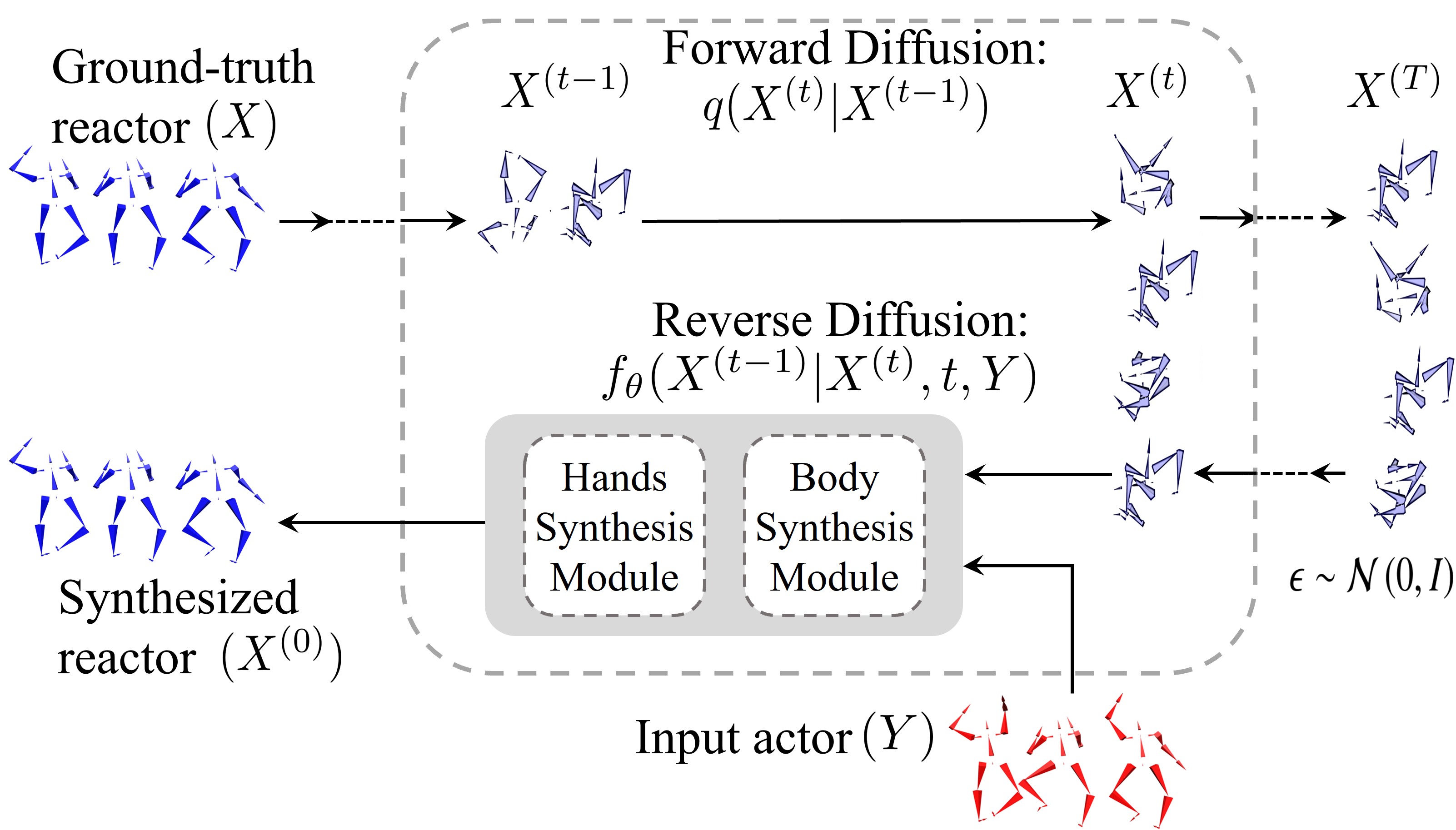}
    \end{minipage}
    \hfill
    \begin{minipage}[c]{0.45\linewidth}
    \centering
        \caption{\textbf{ReMoS Overview.} Given the motion of the actor (\textit{bottom-middle}, in \textitred), we synthesize a plausible motion for the reactor (\textit{bottom-left}, in \textitblue). We achieve this using a denoising diffusion-based probabilistic model (\textit{center}) trained on reactive motion sequences (\textit{top-left}, in \textitblue).}
        \label{fig:overview}
    \end{minipage}
\end{figure}

With this objective in mind, 
we present 
\method, a novel approach for \underline{Re}active \underline{Mo}tion \underline{S}ynthesis with full-body articulations. 
Inspired by the advancements of DDPMs in 
3D human motion synthesis \cite{tanaka2023role, shafir2023human, PoYifanGolyanik2023DiffSTAR},
we develop a DDPM framework (Sec.~\ref{subsec:Diffusion}) with 
a cascaded, two-stage generation strategy tailored to our problem setting 
(Fig.~\ref{fig:overview}).
In the first stage of our diffusion model, we generate the reactive motion for the reactor's body joints conditioned on the actor's body joints. 
In the second stage, we use the synthesized body motions as an additional parameter to generate masking for appropriate hand motions (Sec.~\ref{subsec:framework}). 
We propose a combined spatio-temporal cross-attention (\bodyatt) mapping between the actors' and the reactors' body motion embeddings, that learns the inter-dependencies in their motions without needing additional annotations.
Further, to synthesize hand interactions, we introduce a hand-interaction-aware cross-attention (\handatt) mechanism to ensure the relevant hand joints react to the actors' motions, thus allowing the network to distill localized hand interaction features.
We ensure accurate coordination between the actor and the reactor by using contact-based reaction loss (Sec.~\ref{subsec:training}) and an inference-time guidance function (Sec.~\ref{subsec:guidance}) that improves the plausibility of the body and hand interactions.
To explore reaction synthesis in complex and diverse two-person scenarios, 
we contribute the \textsc{\dataset} dataset consisting of full-body and finger motion sequences for fast-paced swing dance of Lindy Hop~\cite{spring1997swing}, and the martial art technique of Ninjutsu~\cite{cummins2012search} (Sec.~\ref{sec:dataset}).
In summary, our technical contributions are:
\begin{itemize}
    \bulletitem \textbf{\textsc{\method}.} A novel method for reactive 3D human motion synthesis using a cascaded diffusion framework to generate full-body and hand motions. Our framework generates fine-grained reactive motions for complex and dynamic contact-based interactions and derives the reactions  
    directly from the actor's motions, without requiring explicit label or text annotations.
    \bulletitem \textbf{Interaction-Based Attentions.} A combined spatio-temporal cross attention (\bodyatt) mechanism to enforce coherence between the body movements of the actor and the reactor, and 
    a hand-interaction-aware cross-attention (\handatt) mechanism to 
    enforce the appropriate hand-based interactions between the two characters. 
    \bulletitem \textbf{\textsc{\dataset}.} 
    A new dataset for two-person interactions under complex scenarios of Lindy 
    Hop dancing and Ninjutsu.
    The dataset consists of $\sim275.7\mathrm{K}$ motion frames with multiview RGB videos and 3D full-body motion capture of the two interacting persons. It further includes finger-level articulations. 
\end{itemize}
We evaluate our approach in different scenarios, including Lindy Hop dancing, Ninjutsu,  Acrobatics~\cite{guo2022multi} and Kickboxing~\cite{Shum_InHAC}, and report state-of-the-art performance (Sec.~\ref{subsec:evaluation_metrics}).
We also report a user study on comparing the visual quality of \method~compared to the ground truth and the baseline methods (Sec.~\ref{subsec:user_study}).
\section{Related Work} \label{sec:relatedwork}
\paragraph{\textbf{Multi-Person 3D Motion Synthesis.}}
Synthesizing close interactions between two or multiple persons is a challenging task in animation.
Early works in multi-person interaction are based on motion graphs~\cite{shum2007simulating}, interaction patches~\cite{shum2008interaction}, momentum-based inverse kinematics and motion blending~\cite{komura2005animating}, and topologically-based pose representations~\cite{ho2007planning, ho2009character}, to name a few.
The increasing availability of interaction datasets~\cite{yun2012SBU, Shum_InHAC, senecal2020salsa, fieraru2020three, liu2019ntu, kundu2020cross} has led to a rise in data-driven approaches for synthesizing digital partners or opponents in multi-person interaction settings.
Mousas~\cite{mousas2018performance} uses a hidden Markov model to control the dance motions of a digital partner in an immersive setup. 
Ahuja et al.~\cite{ahuja2019react} introduce a residual-attention model to generate body pose in a conversation setting conditioned on two
audio signals and the opposite person's body pose.
Starke et al.~\cite{starke2020local} propose a phase function-based network that learns asynchronous movements of each bone and its interaction with external objects.
Guo et al.~\cite{guo2022multi} propose the Extreme Pose Interaction dataset with a two-stream network with cross-interaction attention modules for forecasting pose sequences of two interacting characters.
GAN-based models~\cite{men2022gan, goel2022interaction} generate motions for an interacting person conditioned on an input character and class labels depicting the type of reaction performed. 
These methods leverage the daily interaction datasets such as SBU Kinect~\cite{yun2012SBU} and 2C~\cite{Shum_InHAC}. 
InterFormer~\cite{Interformer2023} uses an interaction transformer with both spatial and temporal attention
to generate reactive motions given some initial seed poses of both characters.
It can synthesize sparse-level interactions based on the motions of the K3HI~\cite{hu2014two} and the DuetDance~\cite{kundu2020cross} datasets.
Concurrent to our work, Duolando~\cite{siyao2023duolando} uses an off-policy reinforcement learning model to predict tokenized motion for a leader and follower conditioned on music.
The aforementioned methods either depend on seed motion as input or require additional conditions to drive the motion.
In contrast, ReMoS focuses on synthesizing well-synchronized full-body and hand motions for the reactor conditioned only on the actor's 3D motions, without using any additional labels or prompts. 
\paragraph{\textbf{Diffusion Based 3D Motion Synthesis.}}
Denoising diffusion models~\cite{SohlDickstein2015} have recently demonstrated their high potential in generative human motion modeling, specifically in single-person conditional motion synthesis.
Conditional single-person motion generation has been performed using diffusion-based approaches for co-speech gestures~\cite{ao2023gesturediffuclip, zhu2023taming}, audio-driven motion~\cite{tseng2022edge, dabral2022mofusion, zhou2023ude}, and text-driven motion~\cite{zhang2022motiondiffuse, tevet2022human, shafir2023human, yuan2023physdiff} synthesis.
Diffusion-based techniques have also been extensively used for conditional synthesis for human-object interactions~\cite{li2023controllable, Xu_2023_ICCV, li2023object, kulkarni2023nifty}, hand-object interactions~\cite{ye2023affordance, liu2024geneoh}, and human-scene interactions~\cite{huang2023diffusion}.
Guided motion diffusion models such as GMD~\cite{karunratanakul2023gmd} and TraceAndPace~\cite{Rempe_2023_CVPR} incorporate spatial constraints on motion trajectories, to guide the motion towards a goal at inference time.  
OmniControl~\cite{xie2023omnicontrol} proposes spatial and realism guidance to control any joint for diffusion-based human motion generation. 
For two-person interaction synthesis, 
RAIG~\cite{tanaka2023role} proposes a diffusion-based, role-aware approach for two-person interactions, given separate textual descriptions for each person. 
BiGraphDiff~\cite{chopin2023bipartite} uses graph transformer denoising diffusion to learn two-person interaction conditioned on action labels. 
ComMDM~\cite{shafir2023human} uses a communication block between two MDMs~\cite{tevet2022human} to coordinate two-person interaction generation.
ContactGen~\cite{gu2024contactgen} presents a diffusion-based contact prediction module that adaptively estimates potential contact regions between two humans according to the interaction label.
Liang et al.~\cite{liang2023intergen} propose InterGen, a diffusion-based model to synthesize two-person interactions given text descriptions as input conditions. 
They also propose the InterHuman dataset that includes diverse two-person interaction scenarios with rich text annotations. 
Inspired by these recent approaches, we design our proposed method \method~as a DDPM-based 3D motion-conditioned reaction synthesis method, consisting of a cascaded diffusion model with a combined spatio-temporal cross-attention mechanism. This allows \method~to implicitly learn the fine-grained synchronization between two interacting persons from only one of the person's motions, without any need for additional prompts or labels, and synthesize the corresponding motions of the second person.
\method~further synthesizes plausible hand motions for the reactor to incorporate realistic hand-based interactions. 
\section{Reactive Motion Synthesis}\label{sec:method}
We consider a digital, two-person interaction setting where the motions of one character, the \textit{actor}, are known, and we need to synthesize the motions of the other character, the \textit{reactor}.
\method~aims to synthesize the synchronized reactive motions of the reactor in such a setting.
We denote the reactor's motion as $X = \{X_B, X_H\}$, where $X_B \in \mathbb{R}^{N \times J_B\times 3}$ denotes the 3D full-body joint positions with $J_B$ body joints, $X_H \in \mathbb{R}^{N \times J_H\times 3}$ denotes the 3D finger joint positions with $J_H$ finger joints across both hands, and $N$ denotes the number of frames in the motion sequence.
Likewise, we denote the actor's motion as $Y = \{Y_B, Y_H\}$.
Our goal is to model the conditional probability distribution $\mathrm{P}(X | Y)$ from which we can sample plausible reactive motions.
We model this distribution using conditional denoising diffusion probabilistic models (DDPMs), owing to their unique strengths in capturing temporal coherence of motion data and handling complex motion dynamics~\cite{tevet2022human, dabral2022mofusion, tseng2022edge}.
We elaborate our methodology below.
\subsection{DDPM for Reactive Motion Synthesis}
\label{subsec:Diffusion}
The diffusion process consists of two steps: forward or \textit{destructive} diffusion, and reverse or \textit{denoising} diffusion.
In the forward process, we progressively corrupt a clean reactive motion sequence $X$ by adding Gaussian noise $\epsilon$ to it for $T$ steps. 
With sufficiently small noise and large $T$, we can get $X\spsparens{T} \sim \mathcal{N}\parens{0, I}$ following closed-form formulation of Ho et al.~\cite{ho2020denoising}: 
\begin{equation}
    X\spsparens{t} = \sqrt{\bar{\alpha}_t}X\spsparens{0} +\sqrt{1-\bar{\alpha}_t}\mathbf{\epsilon}, \qquad \epsilon \sim\mathcal{N}\parens{0, I},
    \label{eqn:reparameterization}
\end{equation}
where $\bar{\alpha}_t$ controls the rate of diffusion and $t \in \bracks{0, T}$.
Reversing this diffusion process allows for sampling novel motion sequences from a multivariate Gaussian distribution $\mathit{p}\parens{X\spsparens{T}} \sim \mathcal{N}\parens{0, I}$ as
\begin{equation}
     \mathit{p}\parens{X\spsparens{0}} = \mathit{p}\parens{X\spsparens{T}}
    \prod_{t=1}^T \mathit{p}\parens{X\spsparens{t-1} | X\spsparens{t}}. 
    \label{eqn:reverse_diff}
\end{equation} 
We approximate the computationally intractable term $\mathit{p}\parens{X\spsparens{t-1} | X\spsparens{t}}$ with a learnable function $\mathit{f_{\theta}}\parens{X\spsparens{t}, t}$, and optimize 
$\theta$ to encode the space of human reactive motions.
Our reactive motion $X$ is also conditioned on the actor's motion $Y$. Therefore, we modify the learnable function as $X\spsparens{t-1} =\mathit{f}_{\theta}\parens{X\spsparens{t}, t, Y}$.
Following 
recent works 
\cite{ramesh2022hierarchical, tevet2022human}, we estimate the original motion $X\spsparens{0} = \mathit{f}_{\theta}\parens{X\spsparens{t}, t, Y}$ from our diffusion model by iterating through all $t$ denoising 
steps during inference. 

\subsection{\method~Framework}
\label{subsec:framework}
\begin{figure*}[t]
\centering
    \includegraphics[width=1.0\linewidth]{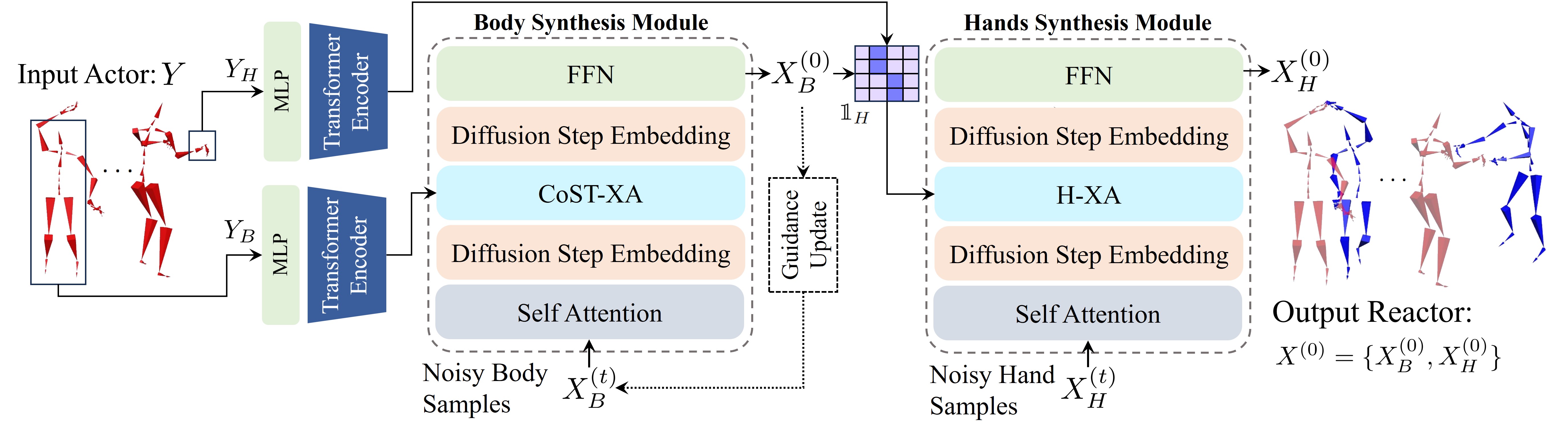}
    \caption{\textbf{\method~Framework.} Given the full-body sequence of the actor (\textit{left}, in \textitred), we input noisy body and hand samples (\textit{from below}) in a cascaded fashion. We synthesize the body samples first, and use them for hand-interaction-aware attention masking (\textit{top-center}) to synthesize the denoised hand samples (\textit{top-right}). The full-body reactive motion is a concatenation of the denoised body and hand samples (\textit{right}, in \textitblue). 
    }
    \label{fig:model_fig}
\end{figure*}

To generate fine-grained reactions with appropriate hand motions, \method~decodes the reactive motion of $X$ in a cascaded fashion (\cref{fig:model_fig}). It first estimates the full-body joints $X_B$, and then the hand joints $X_H$ as
\begin{align}
X_B\spsparens{0} &= \mathit{f}_{{\theta}_B}\parens{X_B\spsparens{t}, t, Y_B}, \label{eqn:body}\\
X_H\spsparens{0} &= \mathit{f}_{{\theta}_H}\parens{X_H\spsparens{t}, t, Y_H, 
\mathbbm{1}_{H_A}\parens{Y_B} \mathbbm{1}_{H_R}\parens{X_B\spsparens{0}}}, \quad\text{and} \label{eqn:hand}\\
X\spsparens{0} &= \braces{X_B\spsparens{0}, X_H\spsparens{0}}\label{eqn:final},
\end{align}
where $\mathbbm{1}_{H_A}$ and $\mathbbm{1}_{H_R}$ are binary, spatio-temporal hand-interaction mask functions that determine which hand joints of the reactor and actor interact at any given frame.
Our cascaded framework comes from the observation that full-body articulations occur at significantly different scales than hand articulations.
Since the body motions affect the hand articulations, our cascaded framework feeds the generated body motions as an additional condition to the hand generation module (Fig.~\ref{fig:model_fig}).
We also train the body and the hand generation modules separately, with additional hand-interaction-aware attention mapping for the hand joints.
Previous work~\cite{ghosh2022imos} used such disjoint training strategy on conditional motion synthesis for single persons. To better accommodate for two-person interactions in our case, we benefit from using a cascaded diffusion strategy~\cite{ho2022cascaded}.  
\paragraph{\textbf{Body Synthesis Module.}}
In the first stage of our cascaded framework, we diffuse the reactor's body motion, $X_B\spsparens{t}$, at each diffusion step $t$, and feed it into a transformer decoder~\cite{vaswani2017attention} block. 
To condition the reactor's body motion on the actor's body motion $Y_B$, we introduce a combined spatio-temporal cross-attention (\bodyatt) mapping.
\bodyatt~combines the spatial and temporal features of the actor's and the reactor's motions through an attention matrix 
to simultaneously learn the inter-dependencies between each pair of reactor and actor body joints $\parens{x_{n,j}, y_{n,j}}$, $x_{n,j} \in X_B\spsparens{t}$ and $y_{n,j} \in Y_B$, respectively, at each frame. 
Denoting the query from the reactor's motion features as 
$Q_B \in \mathbb{R}^{N J_B\times 3}$
and the key-value pair from the actor's motion features as
$K_B \in \mathbb{R}^{N J_B\times 3}$ and $V_B \in \mathbb{R}^{N J_B\times 3}$, respectively,
we formulate an attention matrix of dimensions $N J_B \times N J_B$. 
In contrast, most previous approaches~\cite{tanaka2023role, liang2023intergen} use cross-attention only on the temporal features of the motion using an attention matrix of shape $N \times N$.
InterFormer~\cite{Interformer2023} sequentially uses spatial and temporal cross-attention modules instead of combining them,
which may result in a partial loss of information
on fine-grained, inter-person interactions across time, especially in the absence of additional annotations (prompts or action labels) to condition the reactions.
Our proposed cross-attention module, which we define as 
\begin{align}
\textrm{\bodyatt} &= 
\operatorname{softmax}\parens{\frac{Q_B K_B^T}{\sqrt{d_{K_B}}}} V_B ,
\label{eqn:CoST-XA}
\end{align}
crucially considers combinations of
spatial and temporal interaction features between different body segments of the actor and the reactor to efficiently synchronize their motions. 
We simultaneously project the diffusion timestep $t$ at each denoising step into the transformer block after the attention blocks and use a final layer of fully-connected network with SiLU activations~\cite{elfwing2018sigmoid} and batch normalization~\cite{NEURIPS2018_36072923} to generate the body motions.
The first stage output is a vector $X_B\spsparens{0} \in \mathbb{R}^{N\times J_B \times 3}$ representing the reactor's clean, synthesized body motion.
\par
\paragraph{\textbf{Hands Synthesis Module.}}
The second stage of our cascaded framework synthesizes the reactor's hand joints.
We input noisy samples $X_H\spsparens{t}$ into a similar transformer decoder block conditioned on the actor's hand motions $Y_H$.
Here, we require the attention weights to be high for the hands used in the hand-based interactions and low for the passive hands.
We explicitly enforce this using a spatio-temporal hand-interaction-aware cross-attention mapping (\handatt) to encourage learning localized interaction features.
For \handatt~to work, we introduce binary hand-interaction masks $\mathbbm{1}_{H_A}$ and $\mathbbm{1}_{H_R} \in \mathbb{R}^{N\times J_H}$. Its entries are calculated by thresholding the distances of the actor's and reactor's wrist joints in $Y_B$ and $X_B\spsparens{0}$, respectively, to the nearest body joints of the other person.
The active entries of $\mathbbm{1}_{H_A}$ and $\mathbbm{1}_{H_R}$ determine which of the actor's and reactor's hands are sufficiently close to the body of the other person
(and hence interacting). 
We then project the encoded $Y_H$ features into each decoder layer using \handatt, as
\begin{align}
\textrm{\handatt} = 
\operatorname{softmax}\parens{\frac{\parens{\mathbbm{1}_{H_R} \odot Q_H} \parens{\mathbbm{1}_{H_A}\odot K_H}^T}{\sqrt{d_{K_H}}}} V_H,
\label{eqn:hand-interaction_aware_attn}
\end{align}
where $K_H$ and $V_H$ are the hand motion features from the actor, $Q_H$ are the hand motion features from the reactor with $d_{K_H}$ channels, and 
$\odot$ denotes element-wise product for masking the query and key values.
We project the diffusion timestep $t$ at each denoising step into the transformer block and use fully-connected networks with SiLU activations to generate the clean, synthesized reactor hand motions $X_H\spsparens{0} \in \mathbb{R}^{N\times J_H \times 3}$.
At the end, to obtain the full reactive motion $X\spsparens{0}$, we concatenate $\braces{X_B\spsparens{0}, X_H\spsparens{0}}$ as in Eqn.~\ref{eqn:final}. 
\subsection{Losses and Training Details}
\label{subsec:training}
We train \method~to minimize a weighted sum of three loss terms, the reconstruction loss $\mathcal{L}_c$, the reaction loss $\mathcal{L}_{r}$, and the kinematic loss $\mathcal{L}_{k}$, as
    \begin{equation}
        \mathcal{L} = \lambda_{c}\mathcal{L}_{c} + \lambda_{r}\mathcal{L}_{r} + \lambda_{k}\mathcal{L}_{k},
        \label{eqn:total_loss}
    \end{equation}
where $\lambda_{c}$, $\lambda_{r}$ and $\lambda_{k}$ are scalar weights to balance the individual losses.

\paragraph{\textbf{Reconstruction Loss} $\mathcal{L}_{c}$.} 
This is the standard diffusion data term that minimizes the $\ell_2$-distance between the ground truth and the synthesized reactive motion, as $\mathcal{L}_{c} = \norm{X - X\spsparens{0}}_2$.
While the reconstruction loss provides the vital data term to drive the training, it does not enforce interaction synchronization between the actor and the reactor, or any kinematic constraints on the motion.   

\paragraph{\textbf{Reaction loss} $\mathcal{L}_{r}$.} 
We introduce the reaction loss to ensure accurate timing and spatial positioning of the reactor's motion with respect to the actor's motion. 
We calculate the ground-truth Euclidean distance $d\parens{\cdot, \cdot}$ between each joint of the actor and the reactor at each frame and minimize the deviations from these distances for the actor and the synthesized reactor's motions.
We note that this distance term also implicitly constrains the reactor's joints in frames that are less relevant for interactions, \textit{i.e.}, where the reactor's joints are far from the actor's joints. 
Further, the Euclidean distances for their hand joints in frames with no hand-based contact become drastically high.
To mitigate these concerns, we use an exponentially decaying distance-aware weight $\operatorname{exp}\parens{-d\parens{x_{n, j}, y_{n, j}}}$ at each frame $n$ to focus more on the reactor's joints that are closer to the actor, and therefore more relevant for interaction (see Fig.~\ref{fig:reaction_loss}).
Thus, we get our reaction loss term as
{\small
\begin{align}
    \mathcal{L}_{r} = \frac{1}{NJ}\sum_{n=1}^{N}\sum_{j=1}^{J} \operatorname{exp}\parens{-d\parens{x_{n, j}, y_{n, j}}} \cdot \modulus{d\parens{x_{n, j}, y_{n, j}} - d\parens{x\spsparens{0}_{n, j}, y_{n, j}}}. 
    \label{eqn:IAC_loss}
\end{align}
}

\paragraph{\textbf{Kinematic Loss} $\mathcal{L}_{k}$.} 
To maintain the kinematic plausibility of the generated motions, we
follow existing literature on regularizers for bone length consistency, foot contacts, and temporal consistency~\cite{tseng2022edge, shimada2021neural, shimada2020physcap}.
Our kinematic loss term is a weighted sum of the joint velocity loss $\mathcal{L}_{vel}$, the joint acceleration loss $\mathcal{L}_{acc}$, the bone length consistency loss $\mathcal{L}_{bone}$, and the foot sliding loss $\mathcal{L}_{foot}$, as
\begin{equation}
    \mathcal{L}_{k} = \lambda_v \mathcal{L}_{vel} + \lambda_a \mathcal{L}_{acc} + \lambda_b \mathcal{L}_{bone} + \lambda_f \mathcal{L}_{foot},
    \label{eqn:kin_loss}
\end{equation}
where $\lambda_v$, $\lambda_a$, $\lambda_b$ and $\lambda_f$ are scalar weights (more details in the appendix).
\subsection{Inference Time Spatial Guidance}
\label{subsec:guidance}
While our method synthesizes plausible reactive motions,
it can sometimes spatially misalign the reactor's body to the actor's, espeically at the arm joints, for fast-paced, contact-heavy interactions.
This, in turn, affects finger-joint synthesis in the second stage of our cascaded diffusion.
To improve spatial alignment, we leverage guidance functions~\cite{karunratanakul2023gmd, Rempe_2023_CVPR} that can provide gradients to nudge the sampling process towards a certain direction.
We design a guidance function $G
\parens{\phi, \hat\phi}
\in \mathbb{R}^{N\times J_A \times 3}$,
which minimizes the distance between
the $J_A$ arm joints of the actor ($\phi$) and the $J_A$ arm joints of the synthesized reactor ($\hat\phi$).
Specifically, we re-apply our interaction masks $\mathbbm{1}_{H_A}$ and $\mathbbm{1}_{H_R}$ to determine which of the reactor's hands are more likely to be interacting with the actor, and minimize the distances between the arm joints of the corresponding sides as 
$ G = \argmin_{\hat\phi} \parens{ \norm{\mathbbm{1}_{H_A}\odot \phi - \mathbbm{1}_{H_R} \odot \hat\phi}}$.
We then plug $G$ into the denoising pipeline of our body diffusion module, as
\begin{equation}
    X_B\spsparens{0} = X_B\spsparens{0} - \gamma \nabla_{X_B\spsparens{0}} G\parens{\phi, \hat\phi},\qquad\gamma =\textrm{ guidance scale}.
    \label{eqn:diffusion_gradient}
 \end{equation}
\begin{table}[t]
    \begin{minipage}[c]{0.55\linewidth}
    \centering
    \caption{
    \textbf{ Dataset Comparisons.} Comparing \dataset~with existing multi-person interaction datasets.}
    \label{tab:dataset_comparison}
    \resizebox{\columnwidth}{!}{%
        \begin{tabular}{ l r r c c}
            \toprule
            Dataset & \multicolumn{1}{c}{Motion} & \multicolumn{1}{c}{Duration} & Multi-view & Finger \\
            (Chronologically) & \multicolumn{1}{c}{Frames} & \multicolumn{1}{c}{(hours)} & RGB Videos & Articulation \\
            \midrule
           
            SBU~\cite{yun2012SBU} & $\sim7\mathrm{K}$ & $0.13$ & \xmark & \xmark \\
           
            NTU-26~\cite{liu2019ntu} & $\sim22\mathrm{K}$ & $0.47$ & \cmark & \xmark \\
            
            2C~\cite{Shum_InHAC} & $\sim13\mathrm{K}$ & $0.06$ & \xmark & \xmark \\DanceDB~\cite{senecal2020salsa} & $\sim1.44\mathrm{M}$ & $4.00$ & \xmark & \xmark \\

            DuetDance~\cite{kundu2020cross} & $\sim196\mathrm{K}$ & $1.09$ & \xmark & \xmark \\

            CHI3D~\cite{fieraru2020three} & $\sim486\mathrm{K}$ & $2.70$ & \cmark & \xmark \\

            ExPI~\cite{guo2022multi} & $\sim30\mathrm{K}$ & $0.33$ & \cmark & \xmark \\

             InterHuman~\cite{liang2023intergen} & $\sim107\mathrm{M}$ & $6.56$ & \xmark & \xmark \\

             DD100~\cite{siyao2023duolando} & $\sim200\mathrm{K}$ & $1.92$ & \xmark & \cmark \\
            
            \midrule
            \dataset~(ours) & $\sim275.7\mathrm{K}$ & $2.04$ & \cmark & \cmark \\
            \bottomrule 
        \end{tabular}
    }
    \end{minipage}
    \hfill
    \begin{minipage}[c]{0.4\linewidth}
    \centering
    \includegraphics[width=0.8\linewidth]{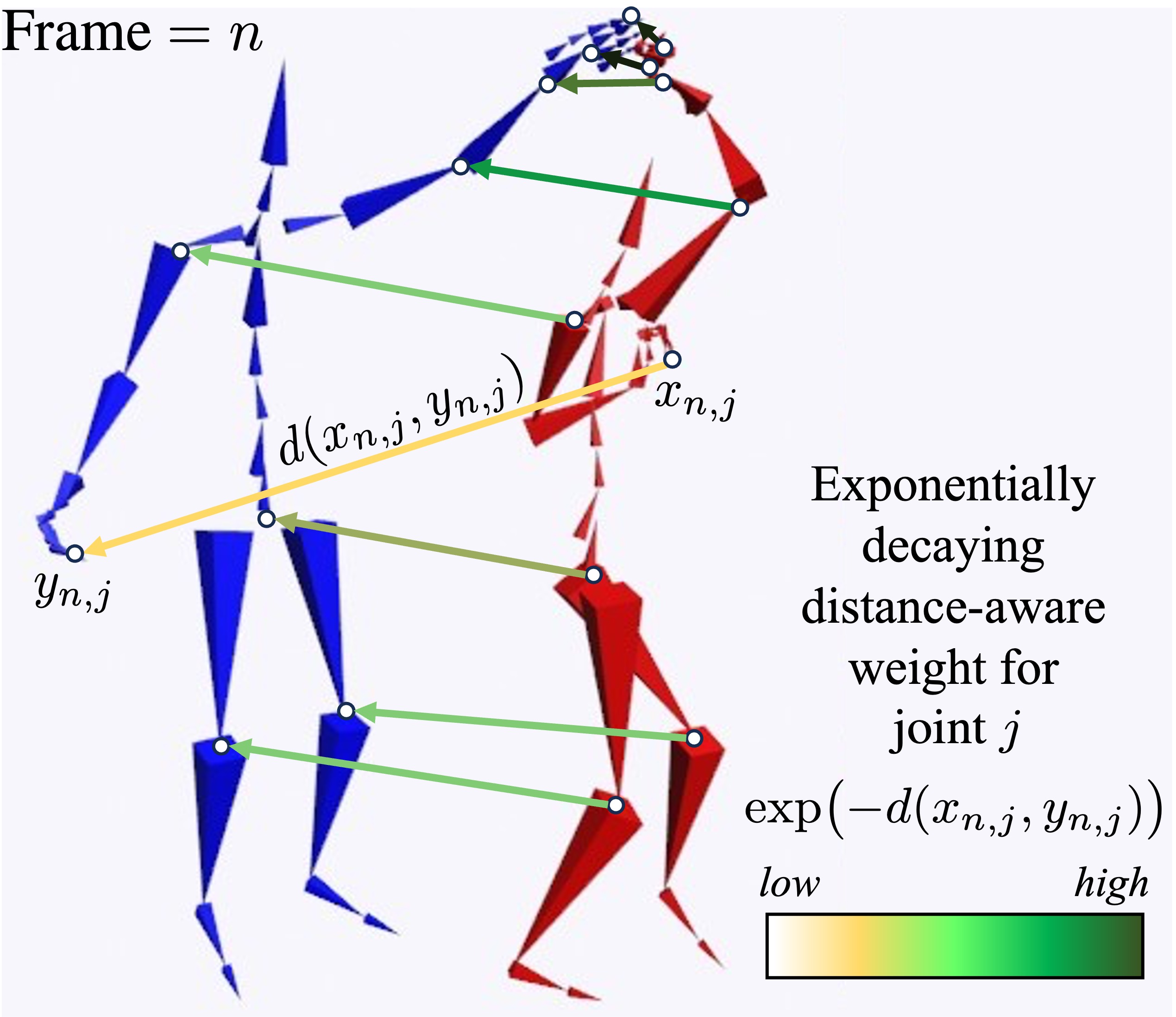}
    \captionof{figure}{\textbf{Visualization of Distance Aware Reaction Loss.} 
    We use an exponentially decaying distance-aware reaction loss to focus more on the reactor’s joints that are closer to the actor.
    }
    \label{fig:reaction_loss}
    \end{minipage}
\end{table}
\section{\dataset~Dataset}\label{sec:dataset}
\par
We propose the \dataset~dataset to facilitate research on contact-based two-person interactions with finger-level articulations.
It contains complex two-person interactions in two separate scenarios: the fast-paced, swing-style Lindy Hop dancing, and the martial arts of Ninjutsu. 
We consider these two types of motions for studying physical interactions for two main reasons. 
First, these motions encompass a significantly \textit{diverse} spectrum of motions
in terms of intensity, speed, and style.
Second, choreographed motion generation is developing as an active field of research and we envision this dataset to facilitate research in interactive applications such as performing in virtual reality~\cite{chan2010virtual} and remote tutoring~\cite{senecal2020salsa}.

\paragraph{\textbf{Data Collection.}}
We invited $4$ trained Lindy Hop dancers and $5$ trained Ninjutsu artists to perform diverse interactions in a multi-view capture studio.
We tracked their motions using a commercially available markerless, multi-view motion-capture system~\cite{Captury2023},
which tracks $93$ degrees-of-freedom driving $69$ body joints.
Capturing motions in a markerless setting enables the performers to move uninhibited, while also making the data suitable for training monocular or multi-view motion capture methods under severe inter-person occlusions.
We also capture RGB videos from $120$ camera views for each sequence. 
The dataset includes 3D skeleton poses with full-body and finger annotations, foreground-background segmentation masks, and 3D surface reconstructions of the subjects. 
We capture sequences of multiple lengths totaling ${\sim}275.7\mathrm{K}$ frames ($2.04$ hours) of motion data from each view at a frame rate of $50$ fps for the Lindy Hop and $25$ fps for the Ninjutsu. 
Out of all the frames, around $150\mathrm{K}$ frames have hand-based interactions between the two characters where the closest distance between the finger joints of the actor and the reactor is within $50$ mm. We present more dataset statistics in the appendix.

\paragraph{\textbf{Dataset Comparison.}}
Table~\ref{tab:dataset_comparison} shows a comprehensive comparison of \dataset~with existing multi-person interaction datasets.
Existing datasets consisting of two-person interactions, such as SBU~\cite{yun2012SBU}, K3HI~\cite{hu2014two}, NTU-26~\cite{liu2019ntu} and 2C~\cite{Shum_InHAC}, are limited in size and motion capture quality. 
They typically feature simple actions, such as handshakes, punching, pushing and kicking, with weak interactions, and do not capture hand motions.
The recent ExPI~\cite{guo2022multi} dataset captures Lindy Hop aerial sequences to model more complex interactions, and the InterHuman~\cite{liang2023intergen} dataset features both daily motions (\textit{e.g.}, passing objects, greeting, communicating) and professional motions (\textit{e.g.}, Taekwondo, Latin dance, boxing). However, these datasets do not provide finger-level motion capture data, which is a key requirement to intricately model inter-human activities. 
Two-person dance datasets, such as DuetDance~\cite{kundu2020cross} and DanceDB~\cite{senecal2020salsa}, also lack the hand motion data needed for modeling interactions. 
Only the concurrent work, Duolando~\cite{siyao2023duolando}\footnote[1]{unpublished at the time of our paper submission} proposes the DD100 dataset with strong interactions between two dancers and provides hand motion data. 
\section{Experiments and Results} \label{sec:experiments}
We conduct comprehensive experiments to evaluate \method~on multiple two-person datasets covering a wide range of interaction scenarios. We perform training and evaluation on multiple large-scale datasets, including 2C~\cite{Shum_InHAC}, ExPI~\cite{guo2022multi}, and our proposed \dataset.
We provide the implementation details, report quantitative and qualitative comparisons on standard evaluation metrics and through a user study, report ablations, and show how to apply ReMoS as a motion editing tool.
We provide details on the dataset preparation in the appendix.
\subsection{Implementation Details}\label{subsec:implementation}
As a pre-processing step, we normalize the two-person body poses by translating the actor's motion $Y$ and the reactor's motion $X$ together, such that the root joint of $Y$ is at the global origin for all $N$ frames.
We then compute the relative 3D joint coordinates of $X_B$ and $Y_B$ w.r.t. the root joint of $Y_B$.
For each hand, we compute the relative 3D joint coordinates of $X_H$ and $Y_H$ w.r.t. the corresponding wrist joints of $X_H$ and $Y_H$.  
Using normalized inter-person, root-relative joint positions benefits the stability and convergence of our model.
\method~uses $T=500$ diffusion steps where $\bar{\alpha}$ changes linearly from $0.0002$ to $0.02$.
We train for about $64\mathrm{K}$ iterations on both sequences of \dataset~using Adam~\cite{adam} with a base learning rate of $10^{-5}$ and a batch size of $64$.
We decay the learning rate using a Step LR scheduler with a step size of $5$ epochs and a decay rate of $0.99$.
We use $d=256$ for our latent embedding representation and use $6$ layers in our transformer decoder with $4$ heads for calculating the attention.
The training takes around $8$ and $11$ hours on \gpu~for Lindy Hop and Ninjutsu, respectively.
The inference time is ${\sim}24s$ to generate $50$ frames of full-body and finger motions ($12.5s$ for body synthesis and then $11.5s$ for hand synthesis).
We set $\lambda_{c}, \lambda_{r}, \lambda_v=10.0$, $\lambda_a, \lambda_k, \lambda_b=1.0$ and $\lambda_f=20.0$ (only applied after $100$ training epochs) as the loss term weights in Eqns.~\ref{eqn:total_loss} and~\ref{eqn:kin_loss}. We set the guidance scale $\gamma = 10^{-3}$ in Eqn.~\ref{eqn:diffusion_gradient}.

\subsection{Baselines and Ablated Versions}
\label{subsec:baselines}
For baselines, we choose the closest motion synthesis methods in a two-person setting, namely, MixNMatch \cite{goel2022interaction}, InterFormer~\cite{Interformer2023}, ComMDM~\cite{shafir2023human}, Role-Aware Interaction Generation (RAIG)~\cite{tanaka2023role}, and InterGen~\cite{liang2023intergen}.  InterFormer~\cite{Interformer2023} was originally trained in a reactive motion synthesis setting without additional annotations, such as action labels or text descriptions, and we maintain this setup. 
For the other methods, we mask out their input text/label embeddings to comply with our annotation-free setting.
We re-train these methods on \dataset~with
a thorough hyper-parameter search and report their best performances. We provide more training details in the appendix.

\begin{table}[t]
    \caption{ \textbf{Quantitative Evaluation on \dataset.} We compare \method~with our baselines and ablated versions (Sec.~\ref{subsec:baselines})
    on the \dataset~dataset.
    We evaluate these methods on metrics such as MPJPE, MPJVE, FID, and Diversity.
    $\downarrow$: lower is better, $\uparrow$: higher is better, $\rightarrow$: values closer to GT are better.
    \textbf{Bold} indicates best. 
    }
    \centering
    \resizebox{1.0\linewidth}{!}{%
        \begin{tabular}{lL{0.3cm}cccccL{0.3cm}ccccc}
        \toprule
        \multirow{2}{*}{Method} && \multicolumn{5}{c}{Lindy Hop} & \multicolumn{5}{c}{Ninjutsu} \\
        \cmidrule{3-7}\cmidrule{9-13}
        && MPJPE  & MPJVE &  FID $\downarrow$ & FID $\downarrow$ & 
        Div  &&
        MPJPE & MPJVE &  FID $\downarrow$ & FID $\downarrow$ & Div \\
        && (mm) $\downarrow$ & (mm) $\downarrow$ & (body)& (hands) & $\rightarrow$ &&
        (mm) $\downarrow$ & (mm) $\downarrow$ & (body)& (hands) & $\rightarrow$\\
        \cmidrule{1-1}\cmidrule{3-7}\cmidrule{9-13}
        GT && - & - &  - & - & $7.57$ && - & - &  - & - & $10.51$ \\
        InterFormer~\cite{Interformer2023} && $66.6$ & $8.26$ & $0.53$ & $0.65$ & $4.54$ && $270.2$ & $3.4$ & $0.57$ & $0.68$ & $6.48$ \\
        MixNMatch~\cite{goel2022interaction} && $70.2$ & $10.3$ & $0.77$ & $0.78$ & $2.48$ && $257.2$ & $5.2$ & $0.74$ & $0.72$ & $4.83$ \\
        ComMDM~\cite{shafir2023human} && $59.4$ & $4.41$ &  $0.32$ & $0.53$ & $7.48$ && $201.2$ & $4.1$ & $0.34$ & $0.58$ & $9.98$ \\
        RAIG~\cite{tanaka2023role} && $71.2$ & $4.32$ & $0.47$ & $0.63$ & $8.45$ && $199.1$ & $5.1$ & $0.21$ & $0.63$ & $10.11$ \\
        InterGen~\cite{liang2023intergen} && $62.6$ & $3.92$ & $0.30$ & $0.61$ & $7.21$ && $172.6$ & $3.9$ & $0.32$ & $0.57$ & $9.98$ \\
        \cmidrule{1-1}\cmidrule{3-7}\cmidrule{9-13}
        \textbf{\method~(ours)} && $\mathbf{40.7}$ &  $\mathbf{2.26}$ & $\mathbf{0.12}$ & $\mathbf{0.26}$ & $\mathbf{7.62}$ && $\mathbf{139.2}$ &  $\mathbf{3.3}$ & $\mathbf{0.16}$ & $\mathbf{0.35}$ & $\mathbf{10.26}$ \\
        w/o diffusion && $72.5$ & $4.91$ & $0.58$ & $0.74$ & $4.04$ && $224.5$ & $4.1$ & $0.52$ & $0.64$ & $6.06$ \\
        w/o cascading && $63.9$ & $4.95$  & $0.51$ & $0.55$ & $7.12$ && $223.6$ & $4.2$  & $0.42$ & $0.75$ & $6.62$ \\
        w/o \bodyatt && $44.2$ & $3.62$ & $0.21 $ & $0.39$ & $7.45 $ && $176.6$ & $3.6$ & $0.27 $ & $0.41$ & $8.91 $ \\
        w/o reaction loss && $44.6$ & $3.51$ & $0.22 $ & $0.38$ & $7.31$ && $144.6$ & $3.7$ & $0.23 $ & $0.39$ & $8.99$\\
        w/o spatial guidance && $41.9$ &  $2.34$ & $\mathbf{0.12}$ & $\mathbf{0.26}$ & $\mathbf{7.62}$ && $139.4$ & $3.4$ & $\mathbf{0.16}$ & $\mathbf{0.35}$ & $\mathbf{10.26}$ \\
        \bottomrule
        \end{tabular}
        \label{tab:ReMoCap_metrics}
        }
\end{table}
\par
We also compare our proposed \method~model with five of its ablated versions: 
\begin{itemize}[leftmargin=*]
    \bulletitem \textbf{w/o diffusion.} 
    Training \method~using a transformer encoder-decoder network without using DDPM strategy. 
    \bulletitem \textbf{w/o cascading.} 
    Training \method~with an integrated transformer for all the joints instead of a cascaded strategy for body and hand generation.
    \bulletitem \textbf{w/o \bodyatt.} Using a spatial followed by a temporal cross-attention mapping 
    instead of the combined spatio-temporal features for body synthesis.
    \bulletitem \textbf{w/o reaction loss.} Removing the reaction loss (Eqn.~\ref{eqn:IAC_loss}) in training.
    \bulletitem \textbf{w/o spatial guidance.} Removing spatial guidance $G_A$ (Sec.~\ref{subsec:guidance}) in inference.
\end{itemize}

\subsection{Quantitative Evaluation}
\label{subsec:evaluation_metrics}
We evaluate \method~on standard evaluation metrics. We measure the temporal consistency deviation from the ground truth 3D pose following~\cite{shimada2020physcap} to report the mean per-joint positional error (MPJPE) and the mean per-frame, per-joint velocity error (MPJVE), both in \textit{mm}, on the synthesized motions.
We measure the Fr\'echet Inception Distance (FID)~\cite{heusel2017gans} to compare the distribution gap between the embedding spaces of the generated and ground-truth motions.
As part of our method focuses on generating hand motions, we measure the FID score of body and hands separately for \dataset.
We also compute the latent variance of the generated motions (Diversity)~\cite{tevet2022human,zhang2022motiondiffuse}.
Table~\ref{tab:ReMoCap_metrics} reports the quantitative evaluation of \method~with its baselines and ablations on \dataset.
\method~achieves state-of-the-art performance for both the Lindy Hop pair-dance setting and the Ninjutsu setting.
We observe that methods using denoising diffusion have higher Diversity compared to transformer-based~\cite{Interformer2023} or GAN-based~\cite{goel2022interaction} methods.
This enforces the variability claims of denoising diffusion models.
We note at least $20\%$ improvement in the MPJPE and around $40\%$ improvement in the FID score for the reactor's body motion when using the proposed \bodyatt~mechanism, confirming its benefit in learning fine-grained, inter-person dependencies across time.
We also note an almost $50\%$ improvement in the FID scores of the reactor's hand motion when using the cascaded strategy and the reaction loss. 
Further, the spatial guidance function fine-tuning improves MPJPE and MPJVE. The MPJPE values are overall higher in Ninjutsu than in Lindy Hop as the trajectory of the reactor is more diverse for Ninjutsu.
We also report the evaluation of \method~with its baselines on the ExPI~\cite{guo2022multi} and the 2C~\cite{Shum_InHAC} datasets in Table~\ref{tab:ExPI_2C}.
These datasets do not provide hand motions, so we only evaluate the reactor's body motions, and report state-of-the-art performance of \method.

\begin{table}[t]
    \begin{minipage}[t]{0.65\linewidth}
    \centering
    \caption{ \textbf{Quantitative Evaluation on the ExPI and 2C datasets}.
    We compare \method~with state-of-the-art motion synthesis methods on the ExPI~\cite{guo2022multi} and 2C datasets~\cite{Shum_InHAC}. $\downarrow$: lower is better, $\uparrow$: higher is better, $\rightarrow$: values closer to GT are better.
   \textbf{Bold} indicates best.
   }
    \label{tab:ExPI_2C}
    \resizebox{\columnwidth}{!}{%
        \begin{tabular}{lL{0.3cm}ccccL{0.3cm}cccc}
        \toprule
        \multirow{2}{*}{Method} && \multicolumn{4}{c}{ ExPI} && \multicolumn{4}{c}{2C} \\
        \cmidrule{3-6}\cmidrule{8-11}
        && MPJPE & MPJVE &  FID $\downarrow$ & Div && MPJPE & MPJVE &  FID $\downarrow$ & Div  \\
        && (mm)  $\downarrow$ &  (mm) $\downarrow$ & (body) & $\rightarrow$ & & (mm)  $\downarrow$ & (mm)  $\downarrow$ & (body) & $\rightarrow$\\
        \cmidrule{1-1}\cmidrule{3-6}\cmidrule{8-11}
        GT && - & - &  -  & $2.01$ && - & - &  - & $2.22$ \\
        InterFormer~\cite{Interformer2023} && $99.1$ & $3.56$ & $0.42$ & $1.31$ && $90.7$ & $5.11$ & $0.52$ & $1.45$\\
        MixNMatch~\cite{goel2022interaction} && $122.4$ & $5.56$ & $0.48$ & $1.18 $ && $62.4$ & $6.01$ &  $0.47$ & $1.24$ \\
        ComMDM~\cite{shafir2023human} && $121.4$ & $5.41$ & $0.45$ & $2.48$ && $69.9$ & $3.34$ & $0.49$ & $2.86$\\
        RAIG~\cite{tanaka2023role} && $131.2$ & $3.96$ & $0.53$ & $2.51$ && $91.6$ & $4.92$ & $0.67$ & $4.45 $ \\
        InterGen~\cite{liang2023intergen} && $100.6$ & $3.91$ & $0.43$ & $2.09$ && $67.6$ & $4.01$ & $0.47$ & $2.91$ \\
        \cmidrule{1-1}\cmidrule{3-6}\cmidrule{8-11}
        \textbf{\method~(ours)} && $\mathbf{97.9}$ & $\mathbf{3.52}$ & $\mathbf{0.41}$ & $\mathbf{1.98}$ && $\mathbf{59.1}$ & $\mathbf{3.33}$ & $\mathbf{0.34}$ & $\mathbf{2.07}$ \\
        \bottomrule
        \end{tabular}
        }
    \end{minipage}
    \hfill
    \begin{minipage}[t]{0.32\linewidth}
    \centering
    \caption{
    \textbf{User Study Results.} Mean scores on a five-point Likert scale (scores $1-5$).
    }
    \label{tab:userstudy}
    \resizebox{\columnwidth}{!}{%
        \begin{tabular}{ l c c c c}
            \toprule
            \multirow{2}{*}{Method} && Motion & &Reaction   \\
            && Quality & & Plausibility  \\
            && $\uparrow$ & & $\uparrow$   \\
            \midrule
            GT && $4.86 \pm 0.54$ & & $4.72 \pm 0.56$  \\
            InterFormer && $2.52 \pm 0.61$ & & $2.28 \pm 0.57$  \\
            MixNMatch && $1.92 \pm 0.71$ & & $2.18 \pm 0.57$  \\
            ComMDM && $3.02 \pm 0.47$ && $3.12 \pm 0.52$\\
            RAIG && $2.83 \pm 0.67$&& $2.48 \pm 0.65$ \\
            InterGen && $3.18 \pm 0.57$ && $3.19 \pm 0.53$ \\
            \midrule
            \textbf{\method} && $\mathbf{3.79 \pm 0.55}$ & & $\mathbf{3.88 \pm 0.54}$ \\
            \bottomrule 
        \end{tabular}
    }
    \end{minipage}
\end{table}
\subsection{User Study }
\label{subsec:user_study}
We evaluate the visual quality of our captured ground truth data and the generated reactive motions through a user study.
We show participants $26$ different interaction sequences across the ground truth, our method, and its baselines.
For each sequence, we randomly show them three methods side-by-side and ask them to rate the 3D motions they observe in terms of \textit{(a)} the reactor's motion quality, \textit{irrespective} of the actor's motion (Motion Quality), and \textit{(b)} the plausibility of the reaction \textit{given} the actor's motion (Reaction Plausibility). 
We ask the participants to rank these motions from `1' (worst score) to `5' (best score) in a five-point Likert scale.
Table~\ref{tab:userstudy} reports the mean scores from the responses of $40$ participants, excluding the responses that did not pass our validation checks.
We notice the ground truth motions in \dataset~achieve the highest score of around $96\%$, indicating that participants perceive our captured ground truth motions to be natural-looking, with realistic interactions.
\method~has the second best ranking of around $78\%$, which is almost $21\%$ higher than the baselines, showing that it is preferred more over the baselines.

\subsection{Qualitative Results and Applications}
\begin{figure*}[t]
    \centering
    \begin{subfigure}[t]{0.8\linewidth}
        \centering
        \includegraphics[width=\linewidth]{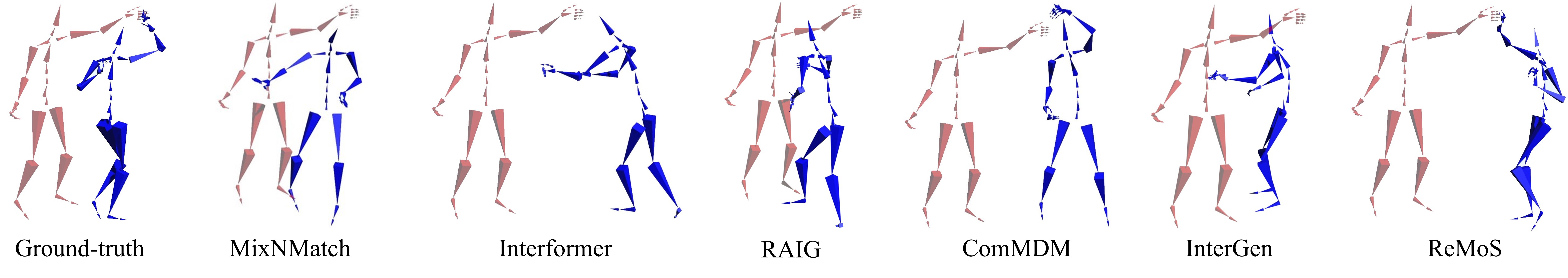}
        \caption{Visual comparison of \method~with GT and baselines.}
        \label{fig:qualitative_comparison}
    \end{subfigure}
    \begin{subfigure}[t]{0.4\linewidth}
        \centering
        \includegraphics[width=\linewidth]{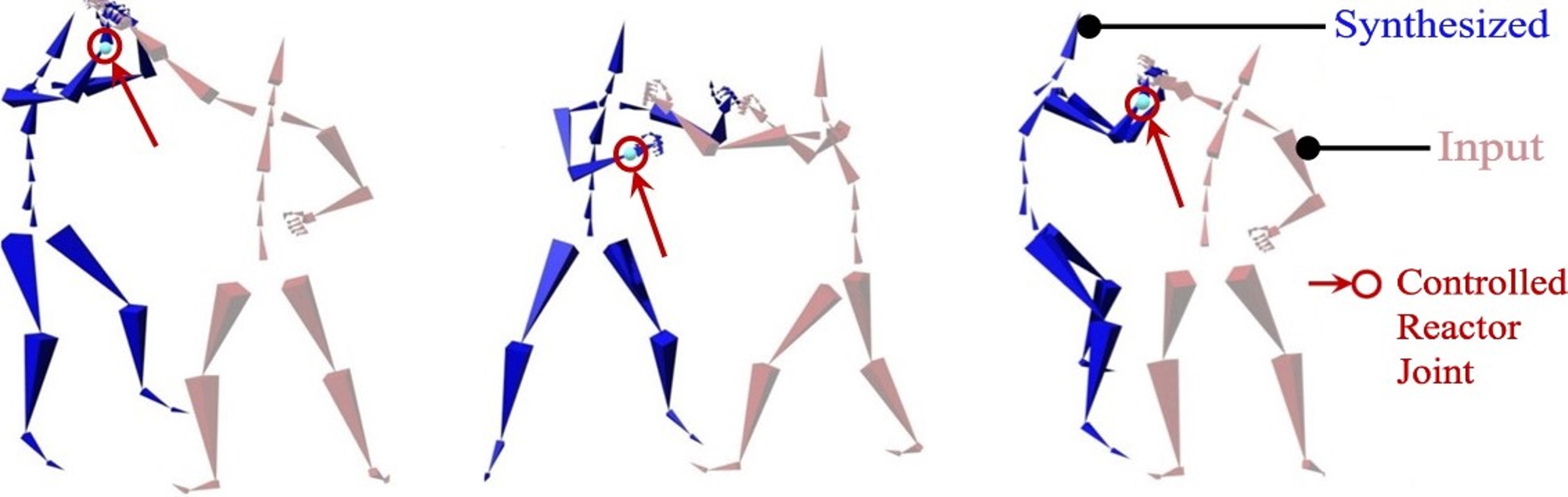}
        \caption{Pose-completion application.}
        \label{fig:appl_pose_completion}
    \end{subfigure}
    \hfill
    \begin{subfigure}[t]{0.4\textwidth}
        \centering
        \includegraphics[width=\linewidth]{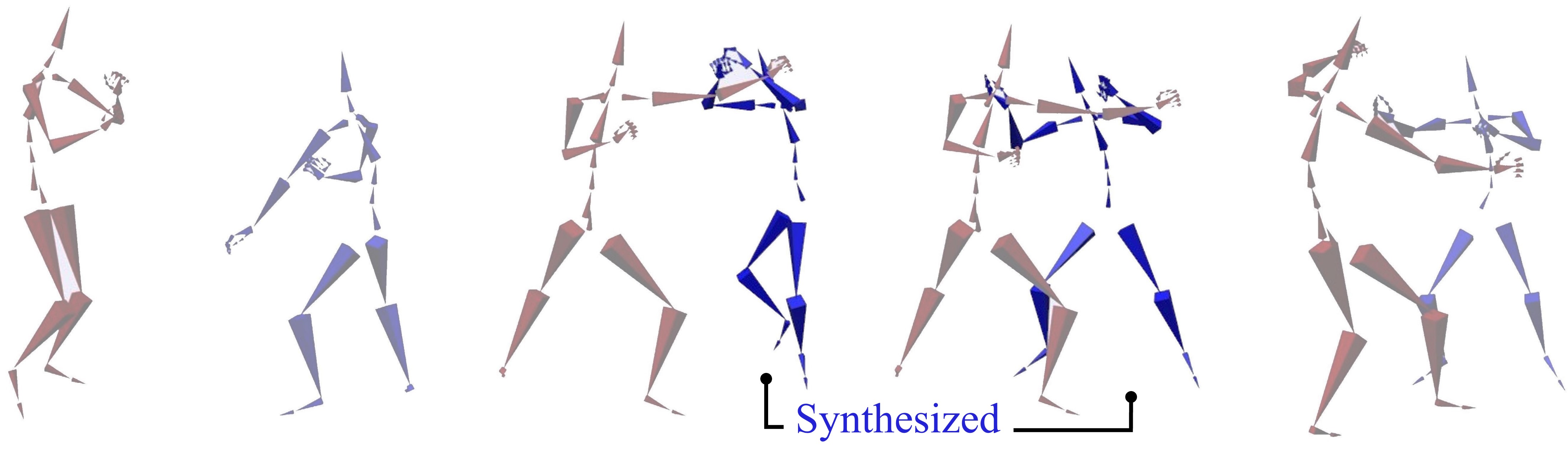}
        \caption{Motion in-betweening application.}
        \label{fig:appl_motion_inbetween}
    \end{subfigure}
    \caption{\textbf{Qualitative Results and Applications.} We show some visual results and the application of \method~as a motion editing tool.
    \textbf{(a)} The reactor (in \textitblue) synthesized by \method~has the most plausible alignment with the actor (in \textitred) compared to the baselines.
    \textbf{(b)} We manually control the right-hand wrist joint of the reactor and let \method~synthesize the remaining body joints conditioned on the actor.
    \textbf{(c)} \method~synthesizes the reactor's motion in-between the start and end frames.
    }
    \label{fig:qualitative_and_appl}
\end{figure*}

Fig.~\ref{fig:teaser} shows qualitative results of \method~on \dataset~and highlights the synthesized hand interactions.
We note that our actor and reactor are \textit{interchangeable}, \textit{i.e.}, depending on the character driving the interaction at a given time, we can swap the actor and the reactor to produce the relevant reactive motions.
Fig.~\ref{fig:qualitative_comparison} shows a visual comparison of \method~with the baselines for one frame of Lindy Hop motion.
\method~synthesizes reaction with the most plausible alignment with the
actor's motion.
We provide detailed visual results in our supplementary video.
Inspired by the applicability of diffusion-based methods for motion editing and controlling, we demonstrate how \method~can also be used as an interactive motion editing tool to control the reactor's motion as desired.
In Figs.~\ref{fig:appl_pose_completion} and~\ref{fig:appl_motion_inbetween}, we show results from two different motion editing applications, namely, \textit{pose completion with controlled joints} and \textit{motion in-betweening}.
We discuss more details in the appendix.

\section{Conclusion}
\label{sec:conclusion} 
\method~brings 3D motion conditioned reaction synthesis to a qualitatively new level by generating diverse, well-synchronized reactions for complex movements and plausible hand motions for contact-based interactions. 
It outperforms the existing baselines both quantitatively and in a user study. 
We also highlight some practical applications of our model, such as pose completion and motion in-betweening, which can lead to the development of useful generative assistants for animators, designers, and creative artists.
Even though we utilize joint positions as parameterization due to their ready availability, we can adapt our approach to accommodate mesh-level inter-person contacts by introducing an offset to the contact threshold between the two bodies, thus simulating skin-to-skin interactions.
We believe there is scope for improvement in the inference speed, where approaches such as Pro-DDPM~\cite{Gandikota_2022_BMVC} and DDIM~\cite{song2020denoising} have shown 
success. 
Future directions also involve scaling the problem towards multi-person motion prediction~\cite{wang2021multi} and considering scene-aware interactions for the characters, which would further enhance immersive user experiences.
%
\section*{Acknowledgements}
This research was supported by the EU Horizon 2020 grant Carousel+ (101017779), and the ERC Consolidator Grant 4DRepLy (770784).
We thank Marc Jahan, Christopher Ruf, Michael Hiery, Thomas Leimkühler and Sascha Huwer for helping with the Ninjutsu data collection, and Noshaba Cheema for helping with the Lindy Hop data collection. We also thank Janis Sprenger and Duarte David for helping with the motion visualizations. 

%
%
\bibliographystyle{splncs04}
\bibliography{remos}
\numberwithin{equation}{section}
\numberwithin{figure}{section}
\numberwithin{table}{section}
\title{\textsc{\method}: 3D Motion-Conditioned Reaction Synthesis for Two-Person Interactions \\ --Appendix--}

\titlerunning{ReMoS -- Appendix}

\author{Anindita Ghosh\inst{1,2,3}\orcidlink{0000-0001-5361-8806} \and
Rishabh Dabral\inst{2,3}\orcidlink{0009-0004-1245-4146} 
\and
Vladislav Golyanik\inst{2,3}\orcidlink{0000-0003-1630-2006} 
\and
Christian Theobalt\inst{2,3}\orcidlink{0000-0001-6104-6625} 
\and
Philipp Slusallek\inst{1,3}\orcidlink{0000-0002-2189-2429}}

\authorrunning{A. Ghosh et al.}

\institute{German Research Center for Artificial Intelligence (DFKI) 
 \and
Max-Planck Institute for Informatics (MPII)
 \and
Saarland Informatics Campus
}

\maketitle
\appendix
We provide additional details on the loss functions used for training \method,
more statistics on the \dataset~dataset, and describe how the datasets and baselines are prepared for evaluation.
We also show some additional results.

\section{Additional Details of Loss Functions}\label{sec:loss}
\paragraph{\textbf{Kinematic Loss Terms.}}
We describe the details of the velocity, acceleration, bone length and foot sliding losses loss terms from Eqn.~10 in the main paper.
To improve the temporal consistency of the motion~\cite{tseng2022edge}, we minimize the joint velocities and joint accelerations between two consecutive frames of the ground-truth reactive motions, $X$, and the synthesized reactive motions, $\hat{X}$, defined as
\begin{small}
\begin{align}
    \mathcal{L}_{vel} &= \frac{1}{N-1}\sum_{n=0}^{ N-1}\norm{\parens{X^{n+1} - X^{n}} -
    \parens{\hat{X}^{n+1} - \hat{X}^{n}}}^2_2, \\
    \mathcal{L}_{acc} &= \frac{1}{N-2}\sum_{n=0}^{ N-2}\norm{\parens{X^{n+2} - 2X^{n+1} + X^{n}} -
    \parens{\hat{X}^{n+2} - 2\hat{X}^{n+1} + \hat{X}^{n}}}^2_2,
    \label{eq:vel_acc_loss}
\end{align}
\end{small}
where $N$ is the total number of frames.

Additionally, we introduce a bone length consistency loss, $\mathcal{L}_{bone}$, to ensure that the synthesized reactor joint positions satisfy the skeleton consistency~\cite{liang2023intergen}. We define this loss as
\begin{equation}
    \mathcal{L}_{bone} = \norm{\mathbf{B}\parens{X} - \mathbf{B}\parens{\hat{X}}}^2_2,
    \label{eq:bone_loss}
\end{equation}
where $\mathbf{B}$ represents the bone lengths in a pre-defined human body kinematic tree. 

Further, foot sliding is a common artifact in motion synthesis~\cite{shimada2021neural,
shimada2020physcap}. We constrain this by ensuring that the toe joint in contact with the ground plane has zero velocity.
We use a binary foot contact loss~\cite{tseng2022edge, tevet2022human} on the foot joints of the synthesized pose to ensure that the output motion does not slide across the ground plane, defined as
\begin{equation}
    \mathcal{L}_{foot} = \frac{1}{N-1}\sum_{n=0}^{ N-1}\norm{\parens{\hat{X}^{n+1} - \hat{X}^{n}} \cdot \hat{\mathbbm{1}}^n_{foot} }^2_2,
    \label{eq:foot_loss}
\end{equation}
where $\hat{\mathbbm{1}}^n_{foot} \in \braces{0,1}$ is the foot-ground contact indicator for the synthesized reactive motion $\hat{X}^n$ at each frame $n$.

\section{\dataset~Dataset Analysis}
Our proposed \dataset~dataset covers two types of motion, namely the Lindy Hop dance and the martial art technique of Ninjutsu (see Sec.~4 in the main paper).

\begin{figure*}[t]
    \centering
    \includegraphics[width=\linewidth]{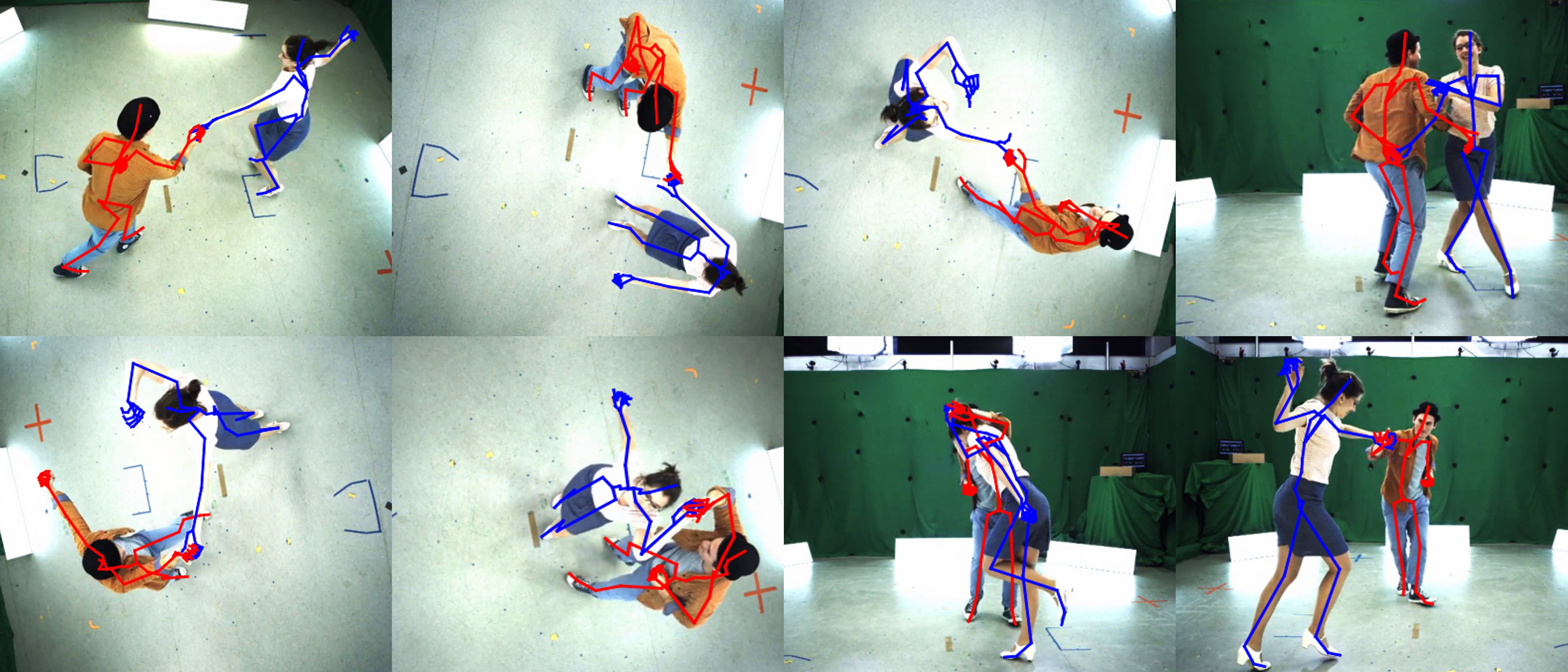}
    \caption{ \textbf{ Samples from the Lindy Hop motion capture for the \dataset~dataset.} We show multi-view RGB samples with corresponding 3D poses from our Lindy Hop motion capture performed by trained dancers. 
    Lindy-hop requires coordination between the two dancers, while also allowing individual dancers the freedom to perform their own motions. This makes it suitable for testing our reactive motion synthesis approach.
    } 
    \label{fig:lindy_hop_captury}
\end{figure*}

\begin{figure*}[t]
    \centering
    \includegraphics[width=\linewidth]{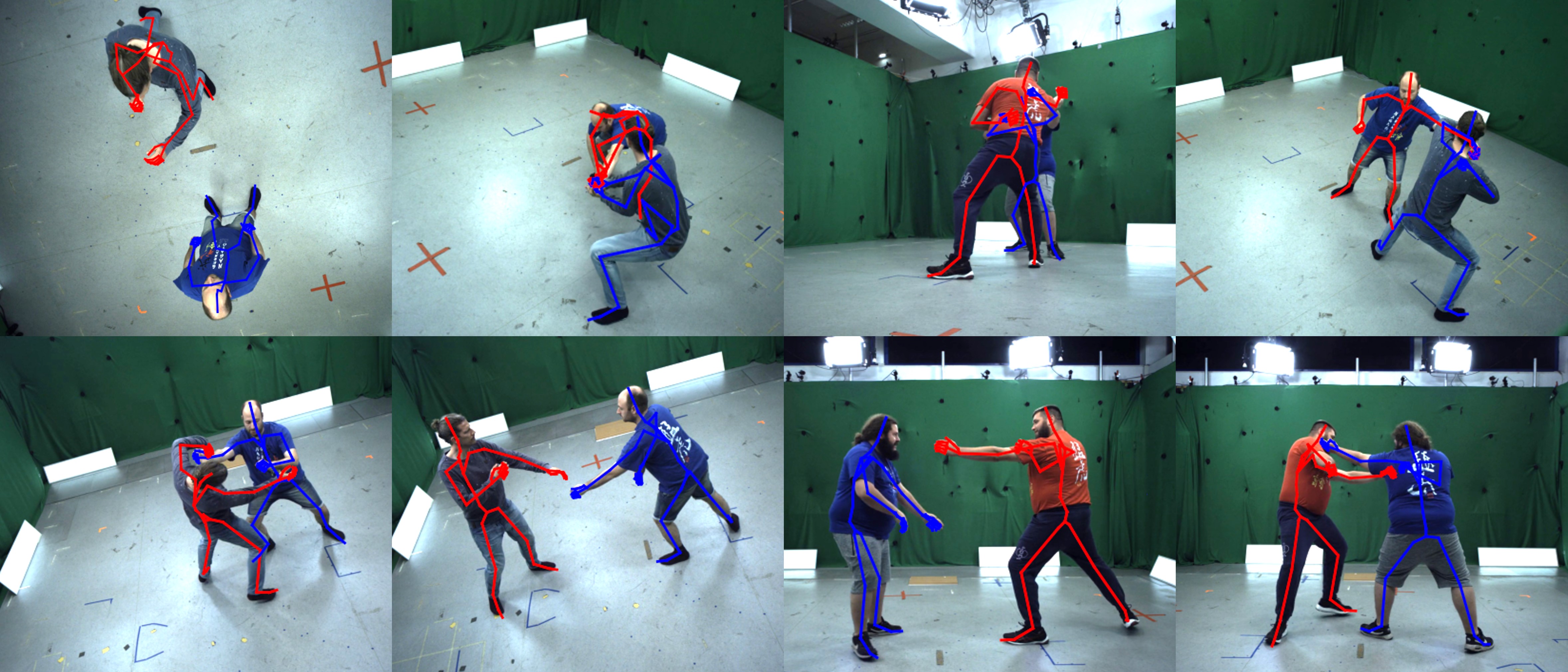}
    \caption{ \textbf{ Samples from the Ninjutsu motion capture for the \dataset~dataset.} We show multi-view RGB samples with corresponding 3D poses from the Ninjutsu motion capture performed by trained artists.
    In contrast to existing martial arts datasets~\cite{Shum_InHAC, zhang2017martial}, we include finger joint motion capture and moves of varying interaction complexity. 
    }
    \label{fig:ninjutsu_captury}
\end{figure*}

\paragraph{\textbf{Lindy Hop motion capture.}}
The Lindy Hop part of the dataset consists of $8$ dance sequences captured at $50$ fps, each around $7.5$ minutes long, resulting in around $174.2\mathrm{K}$ motion frames.
We had $4$ trained dancers, $2$ males (denoted A and B) and $2$ females (denoted C and D), participate in the Lindy Hop motion capture.
We pair the dancers as (A, C), (B, D), (A, D), and (B, C).
Of these pairings, (A, D) contains dance sequences not performed by the other three pairs (in terms of twists and maneuvers).
We also capture multiview RGB videos at $50$ fps from $116$ camera views for each sequence, which can benefit two-person pose reconstruction work in the future.
We show samples from these videos in Fig.~\ref{fig:lindy_hop_captury}.
From these sequences, almost $145.2\mathrm{K}$ frames have a hand-in-hand contact between the two dancers with a contact threshold of $50$ mm between the finger joints of the two dancers.
By increasing the contact threshold to $100$ mm, the number of frames where the two dancers have contact increases to $147.5\mathrm{K}$.
\paragraph{\textbf{Ninjutsu motion capture.}}
The Ninjutsu part of the dataset consists of $79$ sequences each captured at $25$ fps. 
The sequences vary in length with a total number of around $99.8\mathrm{K}$ motion frames resulting in around $66.5$ minutes of motion.
We had $5$ trained, male Ninjutsu artists participate in the Ninjutsu motion capture.
We pair them in all possible combinations and ask them to perform different variations of motion.
Along with the 3D pose, we also capture multiview RGB videos at $25$ fps using $116$ cameras.
We show samples from these videos in Fig.~\ref{fig:ninjutsu_captury}.
From these sequences, almost $81\mathrm{K}$ frames have contact-based interactions between the two performers, where the closest distance between any joints is $50$ mm.
\section{Dataset and Baseline Preparation}
We discuss the preparation of the different datasets and the baseline methods for our evaluation purposes.

\subsection{Dataset Preparation}

\paragraph{\textbf{Preparing the Lindy Hop data in \dataset.}}
We split the dataset such that motions captured from dancer pairs (A, C), (B, D), and (B, C) are in our training set, and motions captured from pair (A, D) are in our test set.
We downsample each motion sequence to $20$ fps and filter the frames where the dancing partners have hand-to-hand contact between the actors' and reactors' finger joints.
We represent each character using $27$ body joints and $22$ hand joints. 
We convert the 3D joint angle representations into joint positions using forward
kinematics and then convert them to root local representations, as explained in Sec.~5.1 in the main paper.
For training, we use a sequence length of $20$ frames.

\paragraph{\textbf{Preparing the Ninjutsu data in \dataset.}}
We divide the whole dataset into roughly $3:1$ train-test ratio and take $28$ sequences of diverse attacking and maneuvering motions for testing, and the rest for training. 
We downsample each motion sequence to $10$ fps, and filter out the frames where the pairs are more than $1$ meter of each other.
We represent each character using $27$ body joints and $22$ hand joints. 
We convert the 3D joint angle representations into joint positions using forward
kinematics and then convert them to root local representations, as explained in Sec.~5.1 in the main paper.
For training, we use a sequence length of $50$ frames. 

\paragraph{\textbf{Preparing the Extreme Pose Interaction (ExPI) Dataset~\cite{guo2022multi}.}}
The ExPI dataset consists of $2$ pairs of professionals performing acrobatics and Lindy Hop aerial sequences.
It consists of $16$ different acrobatic actions.
Each couple consists of a leader and a follower. We aim to synthesize the motions of the followers as they react to the leaders' movements.
We use the \textit{common action split} proposed by the original authors~\cite{guo2022multi}, and split the dataset into train and test sets such that all the actions performed by (A, B) are in the train set and all the actions performed by (C, D) are in the test set.
We represent each subject using $16$ joints (omitting the \textit{`lhead'} and \textit{`rhead'} joints) and convert the global 3D joint positions given in the dataset to root relative joint representations, as explained in Sec.~5.1 in the main paper.
Since the ExPI dataset does not have hand motions, we only train with body motions and forego the hand diffusion stage.
We train \method~for about $20\mathrm{K}$ iterations on the ExPI dataset using the Adam optimizer~\cite{adam}, with a base learning rate of $10^{-5}$ and a batch size of $32$.

\paragraph{\textbf{Preparing the Character-Character Dataset (2C)~\cite{Shum_InHAC}.}}
The 2C dataset consists of full-body motions of kickboxing actions performed by pairs of participants. 
The interactions include motions such as \textit{kicking} and \textit{punching}, with diverse reactions such as \textit{avoiding} and \textit{being hit}. 
We use the pose sequence of the leading character, who throws the punches and kicks, as the acting sequence for our model. 
We aim to synthesize the full body motion of the reacting character, who is blocking or avoiding the moves, as our output.
Following the split of MixNMatch~\cite{goel2022interaction}, we use a roughly $3:1$ train-test ratio to train our method.
Each character contains $25$ joints and we convert the 3D joint angle representations into joint positions using forward kinematics and then convert them to root relative joint position representations, as explained in Sec.~5.1 in the main paper.
Since the 2C dataset does not have hand motions, we only train with body motions and forego the hand diffusion stage.
We train \method~for about $25\mathrm{K}$ iterations on the 2C dataset using the Adam optimizer~\cite{adam}, with a base learning rate of $10^{-5}$ and a batch size of $16$.

\paragraph{\textbf{Preparing the InterHuman Dataset~\cite{liang2023intergen}.}}
We report additional results on the InterHuman dataset in this appendix. It consists of human-human interactions for daily motions, such as passing objects, greeting, and communicating, and professional activities, such as, Taekwondo,
Latin dance, and boxing.
It consists of a total of $7,779$ motions with $22$ joints per person.
We randomly select the pose sequence of one of the characters as the acting sequence for each motion to train our model. 
We aim to synthesize the full body motion of the corresponding other character in each motion as our output.
We follow the split of InterGen~\cite{liang2023intergen} for our experiments. 
Since the InterHuman dataset does not have hand motions, we only train with body motions and forego the hand diffusion stage.
We train \method~for about $45\mathrm{K}$ iterations on the InterHuman dataset using the Adam optimizer~\cite{adam}, with a base learning rate of $10^{-5}$ and a batch size of $64$.

\subsection{Baseline Preparation}
As we mention in Sec.~5.2 in the main paper, we use InterFormer~\cite{Interformer2023},  MixNMatch~\cite{goel2022interaction},  ComMDM~\cite{shafir2023human}, RAIG~\cite{tanaka2023role} and InterGen~\cite{liang2023intergen} as baselines.
We describe how we use each of these methods in our setting.

\paragraph{\textbf{InterFormer~\cite{Interformer2023}.}}
InterFormer consists of a transformer network with temporal and spatial attentions.
It takes an input acting sequence $Y$ and encodes it with spatial and temporal self-attention. 
It also needs the initial pose of the reactor $X$ and predicts the subsequent frames of the reactor in an autoregressive manner. 
It uses information from skeletal adjacency matrices and an interaction distance module that provides information on the interactions.
We use the normalization technique mentioned in Sec.~5.1 in the main paper to normalize the actor's and the reactor's body poses.
We train InterFormer on \gpu~for about $20\mathrm{K}$ iterations for both the LindyHop and the Ninjutsu sets of \dataset, using the Adam optimizer~\cite{adam} with a base learning rate of $10^{-5}$ and a batch size of $128$.
We use $207$ dimensional latent embedding and $6$ layers in the transformer decoder with $3$ heads to calculate the attention.

\paragraph{\textbf{MixNMatch~\cite{goel2022interaction}.}}
MixNMatch proposes an end-to-end framework to synthesize stylized reactive motion informed by multi-hot action labels.
It operates in one of two settings, \textit{interaction mixing} and \textit{interaction matching}.
In \textit{interaction mixing}, it generates a reaction combining different classes of reactive styles according to the multi-label indicator.
In \textit{interaction matching}, it generates the reactive motion corresponding to the interaction type and the input motion.
Our setting is similar to \textit{interaction matching}, where we input the acting sequence into the model and synthesize the reactive sequence.
We mask out the action label defining the interaction type from the input and train the reactor's motion $X$ based on the actor's motion $Y$.
We use the normalization technique mentioned in Sec.~5.1 in the main paper to normalize the actor's and the reactor's body poses.
We train MixNMatch on \gpu~for about $3.6\mathrm{K}$ iterations for both the LindyHop and the Ninjutsu sets of \dataset, using the Adam optimizer~\cite{adam} with a base learning rate of $10^{-5}$ and a batch size of $16$.
We use $256$ LSTM neurons for each spatial slice and $1{,}200$ for the attention layer.

\paragraph{\textbf{ComMDM~\cite{shafir2023human}.}}
ComMDM is proposed as a communication block between two MDMs~\cite{tevet2022human} to coordinate interaction between two persons.
It uses single-person motions from a pre-trained MDM as fixed priors, and a parallel composition with few-shot training that shows how two single-person motions coordinate for interactions.
ComMDM is a single-layer transformer model that inputs the activations coming from the previous layer from the two MDM models, and learns to generate a correction term for the MDM models along with the initial pose of each person.
ComMDM was originally trained for two motion tasks: \textit{prefix completion} and \textit{text-to-motion synthesis}.
We follow the \textit{prefix completion} setting of ComMDM which does not use textual annotations as a condition and was trained to complete $3$ seconds of motion given a $1$ second prefix.
We train ComMDM on \gpu~for about $24\mathrm{K}$ iterations for both the LindyHop and the Ninjutsu sets of \dataset, using the Adam optimizer~\cite{adam} with a base learning rate of $10^{-5}$ and a batch size of $64$.
We use $256$ dimensional latent embedding for the ComMDM block.
During inference, we provide the full ground truth motion of actor $Y$ into the first MDM module.
Thus, the ComMDM block takes in the ground truth features from the first MDM module and the learned features from the second MDM module. In turn,
the output of the second MDM module is the reactive motion $X$ for our setting.

\paragraph{\textbf{RAIG~\cite{tanaka2023role}.}}
Role-Aware Interaction Generation (RAIG) is a diffusion-based model that learns two-person interactions by generating single-person motions for a designated role.
The role is supplied in the form of textual annotations, which are translated into active and passive voices to ensure the text is consistent with each role.
The model generates interactions with two transformers that share parameters, and a cross-attention module connecting them.
The active and passive voice descriptions are proveded as inputs to the corresponding transformers responsible for generating the actor and the reactor.
The transformers consist of cross-attention modules both for language and motion.
To use RAIG as a baseline for our annotation-free setting, we mask out the cross-attention module for the language in both the transformers and train to generate two-person motions unconditionally.
We normalize the interactions as described in the original paper~\cite{tanaka2023role}.
We train RAIG on \gpu~for about $20\mathrm{K}$ iterations for both the LindyHop and the Ninjutsu sets of \dataset, using the Adam optimizer~\cite{adam} with a base learning rate of $2^{-4}$ and a batch size of $32$.
We use $512$ dimensional latent embedding and $8$ attention blocks.
During inference, we freeze the transformer that learns the actor's motion $Y$.
The other transformer generates the reactor's motion $X$, being influenced by the actor's ground truth motion.

\paragraph{\textbf{InterGen~\cite{liang2023intergen}.}}
InterGen is a diffusion-based approach that generates two-person motions from text prompts.
It was originally trained by conditioning on rich textual annotations.
It uses cooperative denoisers with novel weight-sharing and a mutual attention mechanism to improve interactions between two persons.
To use it as a baseline in our annotation-free setting, we mask out the text embeddings from the model input, and train InterGen to generate two-person motions (both actor and reactor) unconditionally.
We use the non-canonical motion representation proposed in the original paper~\cite{liang2023intergen}. 
During inference, we use the customization used in InterGen for \textit{person-to-person generation}.
We take a single-person motion (the actor's motion $Y$) as input, and freeze it during the forward diffusion process.
The frozen weights from the first person propagate into the model, which then uses the ground truth actor's motions to reconstruct the second person's motion (the reactor's motion $X$).
We train InterGen on \gpu~for about $30\mathrm{K}$ iterations for both the LindyHop and the Ninjutsu sets of \dataset, using the Adam optimizer~\cite{adam} with a base learning rate of $10^{-4}$, a cosine LR scheduler, and a batch size of $64$.

\section{Additional Results}
We provide additional results and the trainable parameter counts of all models. We further show how \method~can be used as a motion editing tool for character control applications.

\subsection{Quantitative Evaluation on the InterHuman Dataset~\cite{liang2023intergen}}
We report additional evaluation of \method~compared to its diffusion-based baselines on the InterHuman~\cite{liang2023intergen} dataset in Table~\ref{tab:interhuman_result}.
We report performance on the standard evaluation metrics, including MPJPE, MPJVE, FID, Diversity and Multi-modality. 
For Multi-modality, we generate each sequence $5$ times and report numbers with a $95\%$ confidence interval.
InterHuman dataset does not provide hand motions, so we only evaluate the reactors' body motions. \method~achieves state-of-the-art performance in the aforementioned metrics in the InterHuman dataset, highlighting the utility of our method for diverse forms of two-person interactions.

\begin{table}[t]
    \begin{minipage}[t]{0.65\linewidth}
    \centering
    \caption{\textbf{Quantitative evaluation on the InterHuman dataset~\cite{liang2023intergen}.} We compare \method~with state-of-the-art motion synthesis methods on the InterHuman~\cite{liang2023intergen} dataset. $\downarrow$: lower is better, $\uparrow$: higher is better, $\rightarrow$: values closer to GT are better. \textbf{Bold} indicates best.
    }
    \label{tab:interhuman_result}
    \resizebox{\columnwidth}{!}{%
    \begin{tabular}{lccccc}
        \toprule
        Methods & MPJPE  & MPJVE &  FID & Div &Multi-modality\\
        & (mm) $\downarrow$ & (mm) $\downarrow$ & (body) $\downarrow$ & $\rightarrow$ & $\uparrow$ \\
        \midrule
        GT & $-$ & $-$ & $-$ & $7.74$ & $-$ \\
        ComMDM~\cite{shafir2023human} & $76.4$ & $2.75$  & $0.72$ & $7.17$ & $1.71 \pm 0.5$\\
        RAIG~\cite{tanaka2023role} & $83.2$ & $2.76$ & $0.67$ & $7.26$ & $2.01 \pm 0.6$ \\
        InterGen~\cite{liang2023intergen} & $69.5$ & $2.61$ & $0.59$ & $7.32$ & $2.11 \pm 0.6$ \\
       \midrule
        \method~(ours) & $\mathbf{66.7}$ &  $\mathbf{2.56}$ & $\mathbf{0.56}$ & $\mathbf{7.33}$ & $\mathbf{2.13 \pm 0.3}$\\
        \bottomrule
        \end{tabular}
        }
\end{minipage}
    \hfill
    \begin{minipage}[t]{0.3\linewidth}
    \centering
    \caption{
    \textbf{Trainable parameter counts.}
    }
    \label{tab:params_count}
    \resizebox{\columnwidth}{!}{%
        \begin{tabular}{ l c c }
            \toprule
            \multirow{2}{*}{Method} && Params \\
            && (full model)   \\
            
            \midrule
            InterFormer && $8.2M$  \\
            MixNMatch && $\mathbf{6.5M}$ \\
            ComMDM && $22.2M$\\
            RAIG && $81.2M$ \\
            InterGen && $170M$ \\
            \midrule
            \method~(ours) && $17.4M$ \\
            \bottomrule 
        \end{tabular}
    }
    \end{minipage}
\end{table}
\begin{table}[t]
    \caption{\textbf{Quantitative evaluation on body joints}. We compare the body synthesis module of \method~with state-of-the-art motion synthesis methods on body joints only. \textbf{Bold} indicates the best.}
    \centering
    \resizebox{\columnwidth}{!}{%
        \begin{tabular}{lL{0.02cm}ccccL{0.02cm}cccc}
        \toprule
        Methods && \multicolumn{4}{c}{Lindy Hop (body only)} & \multicolumn{4}{c}{Ninjutsu (body only)} \\
        \cmidrule{3-6}\cmidrule{8-11} 
        && MPJPE $\downarrow$ & MPJVE $\downarrow$ &  FID $\downarrow$ & Div $\rightarrow$  && MPJPE $\downarrow$ & MPJVE $\downarrow$ &  FID $\downarrow$ & Div $\rightarrow$ \\
        \cmidrule{1-1}\cmidrule{3-6}\cmidrule{8-11}
        GT && - & - & - & $7.62$ && - & - & - & $11.5$ \\
        MixNMatch && $69.8$ & $10.5$ & $0.74$ & $2.52$ && $260.1$ & $5.14$ & $0.72$ & $4.94$ \\
        InterFormer && $63.2$ & $8.91$ & $0.52$ & $4.64$ &&  $262.5$ & $3.53$ & $0.51$ & $6.27$ \\
        ComMDM && $50.2$ & $4.42$ & $0.23$ & $7.51$ && $192.4$ & $3.45$ & $0.25$ & $9.83$\\
        RAIG && $68.5$ & $4.01$ & $0.26$ & $9.02$ && $188.3$ & $4.25$ & $0.19$ & $10.14$ \\
        InterGen && $55.1$ & $2.87$ & $0.22$ & $7.49$ && $165.5$ & $3.82$ & $0.23$ & $9.87$ \\
        \midrule
        \method~(ours) && $\mathbf{40.2}$ & $\mathbf{2.21}$ & $\mathbf{0.12}$ & $\mathbf{7.52}$ && $\mathbf{137.2}$ & $\mathbf{3.19}$ & $\mathbf{0.16}$ & $\mathbf{10.26}$ \\
        \bottomrule
        \end{tabular}
        \label{tab:body_metrics}
        }
\end{table}
\subsection{Trainable Parameters}
We report the total number of trainable parameters of \method~as compared to the baseline methods.  Table~\ref{tab:params_count} shows that \method~has lesser trainable parameters than the existing diffusion-based two-person synthesis models~\cite{shafir2023human, tanaka2023role, liang2023intergen}.  
\subsection{Comparison with baselines without hand motions.}

 We compare the body synthesis module of \method~with baselines trained only on the body joints (Table~\ref{tab:body_metrics}). We report state-of-the-art performance for \method~even when finger joints are not included.
\subsection{Motion Editing Applications of \method}
We describe how to use \method~as an interactive motion editing tool, providing control to animators for tasks such as \textit{pose completion} and \textit{motion in-betweening}.
These are crucial applications that are possible due to the strong generative abilities of DDPMs.
We provide visual results of these applications in our supplementary video.

\paragraph{\textbf{Pose Completion with Controlled Joints.}}
When an animator manually customizes some of the reactor's body joints to align with specific animation tasks, \method~can automatically synthesize the reactor's remaining body joints to complete the reactor's motion.
We achieve this by providing the forward-diffused values of the controlled joints as the network input at each diffusion step.
For example, 
to synthesize the motions of the remaining joints of the reactor's body given customized motions for some joints $J_i$ and $J_k$, we set
\begin{equation}
    X_B\spsparens{0} = \mathit{f}_{{\theta}_B}\parens{X_B\spsparens{t}, t, Y_B, \mathbbm{1}_{\braces{J_i, J_k}}},    
\end{equation}
where $\mathbbm{1}_{\braces{J_i, J_k}}$ is a mask we use at each denoising step on all frames to ensure that the joints $J_i$ and $J_k$ are not denoised.
Instead, we populate $J_i$ and $J_k$ with the identical noise vectors as the ones used during forward diffusion, while introducing random noise to the rest of the joints throughout the sequence. 
Thus, animators can incorporate flexible spatial control over chosen joints while \method~synthesizes the remaining joints of the reactor to faithfully capture the interaction.
In Fig.~5b in the main paper, we show the results of a pose-completion application where we manually control the right-hand wrist joint of the reactor and let \method~synthesize the remaining body joints conditioned on the actor.

\paragraph{\textbf{Motion In-Betweening.}}
Likewise, we can use the existing framework to perform motion in-betweening for the reactive sequence. We achieve this by providing some keyframes of the reactive motion and letting \method~synthesize the intermediate frames using a motion in-betweening routine. 
To synthesize the reactor's motion between two given keyframes $N_a$ and $N_b$ through reverse diffusion, we set
\begin{equation}
    X_B\spsparens{0} = \mathit{f}_{{\theta}_B}\parens{X_B\spsparens{t}, t, Y_B, \mathbbm{1}_{\braces{N_a, N_b}}},
\end{equation}
where $\mathbbm{1}_{\braces{N_a, N_b}}$ is a mask we use at each denoising step to ensure that all joints at frames $N_a$ and $N_b$ are not denoised.
Thus, \method~can fill in the motions between the two seed frames as shown in Fig. 5c in the main paper.

\end{document}